\PassOptionsToPackage{svgnames,table,table,dvipsnames}{xcolor}
\PassOptionsToPackage{final}{microtype}

\documentclass[journal]{vgtc}        % final (journal style) without appendices
\onlineid{0}

\vgtccategory{Research}

\vgtcpapertype{evaluation}

\title{A Large-Scale Sensitivity Analysis on Latent Embeddings\\ and Dimensionality Reductions for Text Spatializations}

\shortauthortitle{Atzberger \MakeLowercase{\textit{et al.}}: A Large-Scale Sensitivity Analysis on Latent Embeddings and Dimensionality Reductions for Text Spatializations}
\addtocounter{footnote}{1}
\author{%
  \authororcid{Daniel Atzberger}{0000-0002-5409-7843},
  \authororcid{Tim Cech}{0000-0001-8688-2419},
  \authororcid{Willy Scheibel}{0000-0002-7885-9857},\\
  \authororcid{Jürgen Döllner}{0000-0002-8981-8583},
  \authororcid{Michael Behrisch}{0000-0002-1102-103X}, and
  \authororcid{Tobias Schreck}{0000-0003-0778-8665}
}

\authorfooter{
\item
    Daniel Atzberger, and Willy Scheibel are with University of Potsdam, Digital Engineering Faculty, Hasso Plattner Institute. E-mail: \{firstname.lastname\}@hpi.uni-potsdam.de
\item
    Tim Cech, and Jürgen Döllner are with University of Potsdam, Digital Engineering Faculty. E-mail: \{tcech,rico.richter.1,doellner\}@uni-potsdam.de
\item
    Michael Behrisch is with Utrecht University. E-mail: m.behrisch@uu.nl
\item
    Tobias Schreck is with Graz University of Technology. E-mail: tobias.schreck@cgv.tugraz.at
}

\abstract{%
The semantic similarity between documents of a text corpus can be visualized using map-like metaphors based on two-dimensional scatterplot layouts.
These layouts result from a dimensionality reduction on the document-term matrix or a representation within a latent embedding, including topic models.
Thereby, the resulting layout depends on the input data and hyperparameters of the dimensionality reduction and is therefore affected by changes in them.
Furthermore, the resulting layout is affected by changes in the input data and hyperparameters of the dimensionality reduction.
However, such changes to the layout require additional cognitive efforts from the user.
In this work, we present a sensitivity study that analyzes the stability of these layouts concerning (1) changes in the text corpora, (2) changes in the hyperparameter, and (3) randomness in the initialization.
Our approach has two stages: data measurement and data analysis.
First, we derived layouts for the combination of three text corpora and six text embeddings and a grid-search-inspired hyperparameter selection of the dimensionality reductions.
Afterward, we quantified the similarity of the layouts through ten metrics, concerning local and global structures and class separation.
Second, we analyzed the resulting \numprint{42817} tabular data points in a descriptive statistical analysis.
From this, we derived guidelines for informed decisions on the layout algorithm and highlight specific hyperparameter settings.
We provide our implementation as a Git repository at \href{https://github.com/hpicgs/Topic-Models-and-Dimensionality-Reduction-Sensitivity-Study}{\faGithub~hpicgs/Topic-Models-and-Dimensionality-Reduction-Sensitivity-Study} and results as Zenodo archive at \href{https://doi.org/10.5281/zenodo.12772898}{\textsc{doi}:10.5281/zenodo.12772898}.
}

\graphicspath{{figs/}{figures/}{pictures/}{images/}{./}} % where to search for the images

\usepackage{mathptmx}                  % use matching math font

\DeclareMathAlphabet{\mathcal}{OMS}{cmsy}{m}{n}
\usepackage{numprint}
\npdecimalsign{.} % Default is , (German version)
\usepackage{booktabs}
\usepackage{tabularx}
\usepackage{csquotes}
\usepackage{amsfonts}
\usepackage{amssymb}
\usepackage{extarrows}
\usepackage[export]{adjustbox}
\usepackage{tikz}
\usetikzlibrary{math,calc,positioning,fit,arrows.meta}
\usepackage{makecell}
\usepackage{fontawesome}
\usepackage{calc}

\usepackage{booktabs}
\usepackage{multirow}
\usepackage{diagbox}
\usepackage{collcell}
\usepackage{pgfmath}
\usepackage{pgfplots}
\usepackage{pgfplotstable}

\usepackage{pdfpages}

\pgfplotsset{compat=1.16}

\usetikzlibrary{pgfplots.statistics, pgfplots.colorbrewer}
\usetikzlibrary{matrix}

\usepackage{xcolor}
\usepackage{pifont}% http://ctan.org/pkg/pifont
\usepackage{ifthen}

\newcommand{\CalcCellColor}[1]{%
    \ifthenelse{\equal{\detokenize{#1}}{\detokenize{x}}}
    {}
    {
    \pgfplotsset{colormap/Blues-4}
    \pgfmathfloatparsenumber{#1}%
    \let\value=\pgfmathresult
    \pgfplotscolormapaccess[0:1]{\value}{Blues-4}%
    \xdef\temp{%
        \noexpand\cellcolor[rgb]{\pgfmathresult}%
    }%
    \temp%
    }%
    {#1}
}%

\newcommand{\CalcCellColorInvert}[1]{%
    \ifthenelse{\equal{\detokenize{#1}}{\detokenize{x}}}
    {}
    {
    \pgfplotsset{colormap/Blues-4}
    \pgfmathfloatparsenumber{#1}%
    \let\value=\pgfmathresult
    \pgfplotscolormapaccess[1:0]{\value}{Blues-4}%
    \xdef\temp{%
        \noexpand\cellcolor[rgb]{\pgfmathresult}%
    }%
    \temp%
    }%
    {#1}
}%

\newcolumntype{R}{>{\collectcell{\CalcCellColor}}c<{\endcollectcell}}
\newcolumntype{S}{>{\collectcell{\CalcCellColorInvert}}c<{\endcollectcell}}

\pgfplotstableset{
    /color cells/min/.initial=0,
    /color cells/max/.initial=1000,
    /color cells/textcolor/.initial=,
    color cells/.code={%
        \pgfqkeys{/color cells}{#1}%
        \pgfkeysalso{%
            postproc cell content/.code={%
                \begingroup
                \pgfkeysgetvalue{/pgfplots/table/@preprocessed
                  cell content}\value
                \ifx\value\empty
                    \endgroup
                \else
                \pgfmathfloatparsenumber{\value}%
                \pgfmathfloattofixed{\pgfmathresult}%
                \let\value=\pgfmathresult
                \pgfplotscolormapaccess
                    [\pgfkeysvalueof{/color cells/min}:\pgfkeysvalueof{/color
                      cells/max}]
                    {\value}
                    {\pgfkeysvalueof{/pgfplots/colormap name}}%
                \pgfkeysgetvalue{/pgfplots/table/@cell content}\typesetvalue
                \pgfkeysgetvalue{/color cells/textcolor}\textcolorvalue
                \toks0=\expandafter{\typesetvalue}%
                \xdef\temp{%
                    \noexpand\pgfkeysalso{%
                        @cell content={%
                            \noexpand\cellcolor[rgb]{\pgfmathresult}%
                            \noexpand\definecolor{mapped
                              color}{rgb}{\pgfmathresult}%
                            \ifx\textcolorvalue\empty
                            \else
                                \noexpand\color{\textcolorvalue}%
                            \fi
                            \the\toks0 %
                        }%
                    }%
                }%
                \endgroup
                \temp
                \fi
            }%
        }%
    }
}

\pgfplotstableset{
    colorbarstyle/.style={
        numeric type,
        fixed,
        fixed zerofill,
        precision=1,
        skip 0.=false,
        color cells={min=0,max=1},
        column type=r,
        column type/.add={>{\tiny}}{},
        /pgfplots/colormap name=RdYlBu-CM,
    }
}

\begin{document}

%%%%%%%%%%%%%%%%%%%%%%%%%%%%%%%%%%%%%%%%%%%%%%%%%%%%%%%%%%%%%%%%
%%%%%%%%%%%%%%%%%%%%%% START OF THE PAPER %%%%%%%%%%%%%%%%%%%%%%
%%%%%%%%%%%%%%%%%%%%%%%%%%%%%%%%%%%%%%%%%%%%%%%%%%%%%%%%%%%%%%%%

\firstsection{Introduction}

% Required from template to be in this order
\maketitle

%% \section{Introduction} %for journal use above \firstsection{..} instead
% Motivation
Text data is generated in large amounts from various sources, such as social media platforms, product reviews, news articles, literature, and research articles.
Thereby, text data can be distinguished between single documents, i.e., sequences of words from a vocabulary also known as terms, streams, or sets of documents~\cite{kucher2015text}.
The latter are called text corpora and are usually considered \textit{Document-Term Matrices} (DTMs), which store the absolute term frequencies in the respective documents.
One central question in analyzing a text corpus is providing an overview and displaying the semantic relatedness between the documents~\cite{ward2010interactive}.
Several visualization approaches rely on a two-dimensional scatterplot, where each point represents a document, and the pairwise Euclidean distance between documents reflects their semantic similarity.
Adjacent work in the field augments the two-dimensional scatterplot by utilizing cartographic metaphors, e.g., height fields, icons, or glyphs~\cite{hografer2020state}.
Such scatterplots are derived from a layout algorithm that applies a dimensionality reduction (DR) to the DTM directly or an intermediate latent embedding of the text corpus, which is, for example, derived from a topic model (TM)~\cite{Raval2023ExplainTest}.
This visual representation of a text corpus provides the foundation for a user's \textit{mental map}, i.e., an internal cognitive representation~\cite{beck2017taxonomy}.
If the visualization differs, the mental map is updated, which requires cognitive efforts from the user.

% Problem Statement
Changes to the underlying scatterplot are impeding the information exploration process since \enquote{geometric variations, e.g., rotation, translation, ..., in the projection make the analysis for the user difficult. Internally, the human brain needs to revert these transformations in order to ease the comparison of projections}~\cite{garcia2013stability}.
Consequently, the stability of the visualization mainly depends on the stability of the layout algorithm.
Such stability of a layout algorithm comprises various aspects: 
\begin{enumerate}
    \item[1.] \textit{Stability concerning input data}, i.e., small changes to the input data result in small changes to the layout. This is further known as \textit{Visual-Data Correspondence}~\cite{Kindlmann2014AlgebraicProcess}.

    \item[2.] \textit{Stability concerning hyperparameters}, i.e., small changes to the hyperparameters of the layout algorithm do not change the layout.

    \item[3.] \textit{Stability concerning randomness}, i.e., the layout is not affected by randomness in the initialization.
\end{enumerate}
Furthermore, in the case of time-dependent text corpora, two more aspects are:
\begin{enumerate}
    \item[4.] \textit{Stability concerning corpus size}, i.e., in case of an incrementally growing corpus the positions of previous points are not affected.
    \item[5.] \textit{Stability concerning temporal coherence}, i.e., the changes in the scatterplot capture the evolution of the data points.This is a further aspect of Visual-Data Correspondence.
\end{enumerate}
In presentations of visualization approaches for text corpora, aspects related to the stability of the text layout are rarely considered.
For example, most layouts are derived from \textit{t-distributed Stochastic Neighbor Embedding} (t-SNE), even though it is said to be highly sensitive to its hyperparameters~\cite{wattenberg2016use}.
Only a few studies have considered the stability of DRs, e.g., qualitative studies that inspect scatterplots~\cite{garcia2013stability,bredius2022visual} or theoretical discussions~\cite{Nonato2019MDP}.
Existing quantitative studies focus on two non-text data sets and, therefore, stability of text layout algorithms remains an open gap in the literature~\cite{khoder2012stability,Hamad2018Stability}.

The stability of algorithms and models can be analyzed in a sensitivity analysis, i.e., a quantitative study that analyzes the relative importance of input factors to the output~\cite{saltelli2004sensitivity}.
In this work, we present a sensitivity analysis of layout algorithms for text corpora concerning changes to the input data, hyperparameters, and randomness.
Our approach has two steps: (1) the data generation step and (2) the data analysis step.
In the first step, we derive tabular datasets by measuring aspects related to local, global, and perceptual similarity between selected pairs of scatterplots.
In total, we consider \numprint{38941} scatterplots that are generated from three text corpora, six text embeddings, and four DRs by a grid-search-inspired hyperparameter selection of the DRs.
The scatterplots were created on a computation cluster using 50 nodes and an overall computation time of \numprint[hours]{50000}.
In the second step, we examine the similarities in a descriptive analysis, including a correlation analysis, statistical tests, and by visualizing the data distributions.
Our study makes the following contributions to the field of text visualization and empirical analysis of layout algorithms:
\begin{enumerate}
    \setlength{\itemsep}{0pt}
    \item A set of metrics that quantify the preservation of local and global structures, as well as cluster separation between two scatterplots with the same number of points.
    \item Tabular datasets that capture the similarity between \numprint{42817} pairs of scatterplots, that are derived from three text corpora by applying six embeddings and four DRs.
    \item An analysis of the results concerning the stability of the layout algorithms and guidelines for their effective use.
    \item The implementation and results of the entire pipeline provided as a git repository\footnote{\href{https://github.com/hpicgs/Topic-Models-and-Dimensionality-Reduction-Sensitivity-Study}{\faGithub~hpicgs/Topic-Models-and-Dimensionality-Reduction-Sensitivity-Study}}.
\end{enumerate}

\section{Related Work}
\label{section: related work}

We consider the following three aspects as related work: 
(1) previous studies that focused on the stability of DRs,
(2) comparisons of DRs that propose guidelines for their effective use,
and (3) studies that focus on the visual perception of scatterplots and allow for comparison of scatterplots using similarity measures.

\subsection{Stability of Dimensionality Reductions}
We further distinguish discussions on the stability of DRs into mathematical discussions, qualitative studies, and quantitative analyses.
Nonato and Aupetit presented a survey on DRs, including a discussion on stability concerning input data and their capabilities to map new data points based on their mathematical internals~\cite{Nonato2019MDP}.
Garc{\'\i}a-Fern{\'a}ndez et al. compared six DRs concerning the stability in the input data and hyperparameters by visually comparing their results on six datasets ~\cite{garcia2013stability}.
Similarly, Bredius et al. studied the stability of neural network projections - trained to approximate a DR - concerning changes in the input data~\cite{bredius2022visual}.
Reinbold et al. applied k-order Voronoi diagrams to visualize the neighborhood preservation of several two-dimensional scatterplot representations of a high-dimensional dataset to analyze the stability of MDS and t-SNE concerning input data and hyperparameters~\cite{reinbold2020visualizing}.
Complimentary to qualitative studies, Khoder et al. presented a quantitative study on the stability of DRs regarding input data for hyperspectral images~\cite{khoder2012stability}.
However, their metrics are specific to their domain and can not be adapted in our case.
The most similar method to our work was applied by Hamad et al., who studied the stability of t-SNE by comparing sequences of scatterplots derived from smart home data~\cite{Hamad2018Stability}.
Their quantification of stability relies on the Procrustes distance and a metric that captures neighborhood preservation.
However, we use ten metrics for a more fine-granular quantification of similarity, four DRs, and six latent embeddings to address the specific domain of text corpora visualization.

% Hier suchen wir noch nach einem Übergang
In several application contexts, data is often only incrementally available, e.g., social media or sensor data. 
The capability of a DR to handle such data streams is called \textit{out-of-sample capability}~\cite{Espadoto2021Toward}.
In order to preserve the mental map, the positions of previous points should not be affected by incoming data. Our terminology refers to this aspect as stability concerning corpus size.
Existing evaluations of out-of-sample techniques focus on the overall distance and neighborhood preservation~\cite{Neves2022Fast,Xia2024ParallelFramework} or runtime performance~\cite{Neves2022Fast,Fujiwara2020IncrementalDR,Xia2024ParallelFramework}.
Xia et al. furthermore evaluated the stability of the layout using four metrics~\cite{Xia2024ParallelFramework}.
An overview of specialized out-of-sample techniques is presented by Neves et al.~\cite{Neves2022Fast}.
In order to preserve the mental map, the positions of previous points should not be affected by incoming data.
%
% Temporal Stability
An evolution of data points over time is another form of temporal dependency (stability concerning temporal coherence).
To visualize such evolution, static methods can be adapted, e.g., by using control points known as landmarks~\cite{Nonato2019MDP}.
Furthermore, specialized DRs have been developed.
For example, Rauber et al. presented a variant of t-SNE, where the loss function is adopted to consider temporal coherence~\cite{rauber2016visualizing}.
Vernier et al. presented a quantitative framework for evaluating the temporal coherence of DRs by measuring the preservation of local and global structures~\cite{vernier2020quantitative}, and used it to evaluate the quality of two novel variants of t-SNE~\cite{vernier2021guided}.
Our work differs from previous work, as we specifically analyze the impact of text embeddings on the static stability of two-dimensional layouts.

\subsection{Selecting Dimensionality Reductions}
Besides stability, the capability of a DR to preserve local and global structures in a lower-dimensional representation, the so-called accuracy, is another quality aspect~\cite{Nonato2019MDP}.
Several studies analyzed DRs concerning their accuracy to derive guidelines for their effective use, e.g., Fodor who presented a qualitative comparison of linear and non-linear DRs~\cite{fodor2002survey}.
Further studies that focus on the mathematical principles of DRs were presented by Engel et al., who reviewed DRs from a visualization point of view~\cite{engel2012survey}, Gisbrecht and Hammer, who reviewed non-linear DRs~\cite{gisbrecht2015data}, and Cunningham and Ghahramani, who presented a survey on linear DRs~\cite{cunningham2015linear}.
Complimentary to these theoretical discussions, the accuracy of DRs was evaluated in quantitative studies by using quality metrics~\cite{behrisch2018quality}.
van der Maaten et al. presented an early quantitative discussion on DRs~\cite{van2009dimensionality}.
In their study, the authors compared the accuracy of eleven non-linear DRs and PCA on five synthetic and five real-world datasets.
Gove et al. presented a study on the influence of selected t-SNE hyperparameters and further presented a neural network to recommend better default settings for a given dataset~\cite{Gove2022Guidance}.
Espadoto et al. presented a benchmark that is set up of 18 data sets, 44 DRs, and seven accuracy metrics~\cite{Espadoto2021Toward}.
With this broad sampling, the study has particular value for practitioners.
Atzberger and Cech et al. followed their approach, focusing on text corpora and TMs~\cite{acstrds2023-evaluation-tm-dr, acsds2024-topic-model-influence}.
Their studies are based on benchmarks comprising a set of text corpora, layout algorithms that are combinations of text embeddings and DRs, as well as metrics for quantifying the accuracy and cluster separation.
In an analysis that comprises more than 40k layouts, the authors showed that TMs improve the accuracy of text layouts.
Our work follows the methodology of such quantitative studies based on a benchmark comprising a set of text corpora, layout algorithms, and metrics.
However, even though we leverage parts of the implementation provided~\cite{acstrds2023-evaluation-tm-dr}, our study is concerned with stability which is a different objective than accuracy and thus requires the use of different metrics.

In addition to accuracy, the suitability of a DR for specific tasks needs to be considered too.
Etemadpour et al. compared the performance of four DRs in a user study to support abstract analytics tasks, such as cluster identification~\cite{Etemadpour2015Perception}.
Their results showed that the optimal DR depends on the question and the \enquote{nature} of the data, e.g., whether the data is a text corpus or a set of images.
Xia et al. formulated similar findings after conducting a user study to investigate which DRs are suitable for visual cluster analysis tasks, e.g., cluster identification, membership identification, distance comparison, and density comparison~\cite{Xia2022RevisitingDimensionalityReduction}.
For visual class separation, Wang et al. developed a novel supervised linear DR, which aims to minimize a cost function that relies on class separation metrics~\cite{Wang2018SupervisedDR}.
Furthermore, Morariu et al. presented a model to predict human preferences based on scagnostics, cluster separability metrics, and accuracy metrics~\cite{Morariu2023PredictingUserPreferences}.

\begin{table*}
	\caption{Characteristics for the three datasets containing the number of documents $N$, the size of the vocabulary $n$ after preprocessing, the median size of the documents $l$, the number of categories $k$, and the number of topics $K$ specified for the TMs, as well as the sparsity ratio $\gamma = 1 - u/Nn$, where $u$ denotes the number of non-zero entries in the DTM.}%
	%
%\tiny
\footnotesize%
\centering%
\setlength{\tabcolsep}{8.0pt}%
\renewcommand{\arraystretch}{1.00}%
\vspace{-0.7\baselineskip}
\begin{tabular}{rlrrrrrr}
	\toprule
	\multicolumn{1}{c}{\textbf{Dataset}} & \multicolumn{1}{c}{\textbf{Source}} & \multicolumn{1}{c}{\textbf{$N$}} & \multicolumn{1}{c}{\textbf{$n$}} &\multicolumn{1}{c}{$l$} & \multicolumn{1}{c}{\textbf{$k$}} & \multicolumn{1}{c}{\textbf{$K$}} & \multicolumn{1}{c}{$\gamma$} \\ \midrule
	20 Newsgroup & \href{https://scikit-learn.org/0.19/datasets/twenty\_newsgroups.html}{scikit-learn.org/0.19/datasets/twenty\_newsgroups.html} & \numprint{18846} & \numprint{72370} & 35 & 20 & 20 & 0.9993\\
	Lyrics& \href{https://www.kaggle.com/datasets/karnikakapoor/lyrics}{kaggle.com/datasets/karnikakapoor/lyrics}  & \numprint{10995} & \numprint{32758} & 135 & 4 & 12 & 0.9974\\
	Seven Categories& \href{https://www.kaggle.com/datasets/deepak711/4-subject-data-text-classification}{kaggle.com/datasets/deepak711/4-subject-data-text-classification}  & \numprint{3142} & \numprint{34947} & 198 & 7 & 14 & 0.9962\\
	\bottomrule
\end{tabular}%
    \label{tab:datasets}
\end{table*}

\subsection{Visual Perception and Similarity Measures}
Our analysis of the stability of text layouts is based on a quantification of similarity between scatterplots.
Various metrics have been proposed to describe geometric structures in scatterplots, e.g., the class separation~\cite{Sedlmair2012Taxonomy}, and thus allow for a comparison.
Among the most popular metrics are the scagnostic measures, which result from graph-theoretical characteristics~\cite{wilkinson2005Scagnostics}.
However, user studies have shown that scagnostics are not aligned with human expectations for describing perceptual similarity~\cite{DBLP:conf/chi/PandeyKFBB16,Wang2020RobustScagnostics}.
In addition, more complex models have been developed, which learn abstract representations of scatterplots based on human-labeled data.
For example, Quadri et al. proposed a ranking model for optimizing designs of scatterplots for cluster identification~\cite{Quadri2023AutomaticScatterplotDesign}.
Jeon et al. developed a regression model to estimate cluster ambiguity~\cite{Jeon2024Clams}.
Ma et al. proposed a convolutional neural network to learn a representation of scatterplots for the quantification of similarity~\cite{Ma2020Scatternet}. 
Similarly, Xia et al. presented another convolutional neural network for modeling human perception of visual clusters~\cite{Xia2021VisualClustering}.
Those neural network approaches learn internal representations of scatterplots and allow for a comparison.
However, the individual components have no semantic meaning and thus allow for no further reasoning.
Lehmann and Theisel introduced an equivalence relation on two-dimensional scatterplots based on affine transformations~\cite{Lehmann2016OptimalProjections}.
By selecting a represent of each equivalence class, the number of scatterplots within a scatterplot matrix is reduced.
In particular, this approach does not rely on any further metrics but allows for comparison.

\section{Data Measurement}
\label{section: data generation}

To analyze the stability of layout algorithms for text corpora, we selected three raw text corpora and defined and applied a suitable data processing pipeline.
It is set up of four stages: (1) preprocessing and perturbations, (2) mapping into a latent space using text embeddings, (3) dimensionality reduction to the two-dimensional plane, and (4) pairwise comparison using similarity metrics.

\begin{figure*}[t]
	\centering
	\includegraphics[]{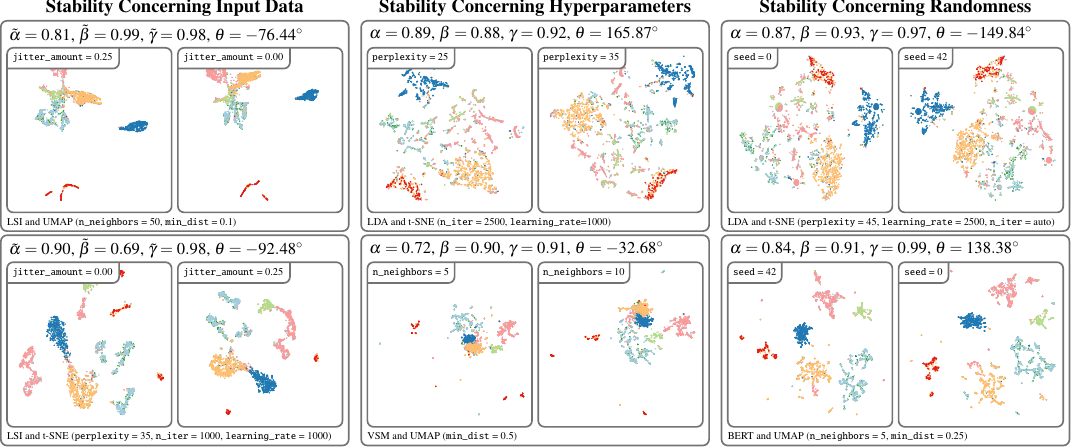}%
	\vspace{-0.5\baselineskip}
	\caption{Exemplary comparison of pairs of scatterplots.
        To analyze the stability concerning input data, we compare pairs of scatterplots that only differ in the amount of jitter applied to the DTM.
        To analyze the stability concerning hyperparameters, we compare pairs of scatterplots that differ in one hyperparameter setting with consecutive values.
        To analyze stability concerning randomness, we compare two layouts that only differ in their seeds.}
	\label{fig:comparisons}
\end{figure*}

\subsection{Preprocessing \& Perturbation}

We selected the following three corpora for our study: \textit{20 Newsgroup}, \textit{Lyrics}, and \textit{Seven Categories}.
All documents of the corpora are written in the English language.
The characteristics, as well as sources, of the text corpora are summarized in \Cref{tab:datasets}.
We preprocessed each corpus as follows:
In the first step, the documents of a corpus are tokenized, i.e., documents are split at whitespaces and stored as lists of terms.
To remove terms that have no semantic meaning and reduce the vocabulary size, we removed all stopwords from the English language and lemmatized the vocabulary.
In addition, we carried out corpus-specific preprocessing steps, such as the removal of email headers in the 20 Newsgroup corpus.
For all details regarding the preprocessing, we refer to our git repository.
After preprocessing, each corpus is represented by a DTM, i.e., each corpus is described by an $Nxn$-matrix, where $N$ denotes the number of documents and $n$ the vocabulary size.
The entry in cell $(i,j)$ gives the frequency of the $j^{\text{th}}$ term in the $i^{\text{th}}$ document. 
Furthermore, each document within a corpus is assigned one unique category that reflects its semantics, e.g., each document within the Seven Categories corpus is associated to either \textit{Computer Science}, \textit{History}, \textit{Maths}, \textit{Accounts}, \textit{Physics}, \textit{Biology}, or \textit{Geography}.
We selected these corpora as their categories are easily interpretable, and the extracted topics show a connection to these categories.
Furthermore, the three corpora vary in the number of documents, vocabulary size, and categories.

To analyze the stability concerning changes in the input data, we apply synthetic perturbations to the DTMs.
We define a jittering perturbation, which adds noise to the DTMs.
Specifically, the entry in cell $(i,j)$ is replaced by:
\begin{align}
    \text{DTM}[i,j] = \max\{0, \text{round}(\text{DTM}[i,j] \cdot (1 + \epsilon[i,j]))\},
\end{align}
where the matrix $\epsilon$ is a $n\times m$-matrix whose entries are randomly drawn from a uniform distribution on the interval $[-\lambda, \lambda]$ and $\lambda \in [0,1]$ controls the amount of jittering.
We chose a relatively simple jittering function to avoid making any further assumptions.

\subsection{Document Embeddings}

Starting from the DTM representation, we further consider five additional document embeddings that are commonly used for corpus visualization and text analytics tasks, resulting in six latent representations for further analysis.
From the perspective of the DTM representation, each document is represented as an $n$-dimensional vector containing the absolute frequencies of individual terms.
This representation, along with the \textit{cosine similarity}, constitutes the \textit{Vector Space Model} (VSM)~\cite{crain2012dimensionality}.
However, the DTM solely accounts for the absolute frequencies of terms within documents, disregarding whether these terms are prevalent across other documents in the corpus.
Often, terms found in only a few documents indicate underlying concepts and hold significant relevance. 
By incorporating the \textit{term frequency-inverse document frequency} (tf-idf) scheme, the VSM can be adapted to address this issue~\cite{aggarwal2012surveyClassification}.
Specifically, the tf-idf of a term $w$ in document $d$ is given by the product of the term-frequency of the term $w$ in $d$ and the inverse document frequency of $w$ in $d$, i.e.,
\begin{align}
    \text{tf-idf}(w,d)
    = \dfrac{n(w,d)}{\sum \limits_{d' \in C} n(w, d')} \cdot \log \Bigg ( \dfrac{\vert C \vert}{\vert \{d' \in C\vert w \in d'\} \vert} \Bigg),
\end{align}
where $n(w,d)$ denotes the frequency of term $w$ in document $d$, and $\mathcal{C}$ denotes the corpus.
The DTM is usually a sparse matrix, i.e., most entries are zero due to documents containing only a fraction of the entire vocabulary.
This observation is exemplified in \Cref{tab:datasets}, where the median lengths are significantly smaller than the vocabulary sizes.

TMs aim to provide a compressed representation of the DTM by grouping co-occurring words into topics.
Topics are represented as vectors of size $n$, where the $i^{\text{th}}$ entry reflects the influence of term $w_i$ within the topic.
Such representations often allow for the inference of human-interpretable concepts.
For instance, \textit{Latent Semantic Indexing} (LSI) employs \textit{Singular Value Decomposition} (SVD) to decompose the DTM or its tf-idf weighted variant into a document-topic matrix and a topic-term matrix~\cite{deerwester1990indexing}.
\textit{Non-Negative Matrix Factorization} (NMF) approximates the DTM or its tf-idf weighted variant as a product of two matrices~\cite{lee1999learning}.
In the case of LSI and NMF, the documents are compared using the cosine similarity.
\textit{Latent Dirichlet Allocation} (LDA) is a probabilistic TM widely used in the visualization domain. 
LDA assumes a generative process underlying a corpus, resulting in topics represented as multinomial distributions over the vocabulary and documents represented as multinomial distributions over topics~\cite{blei2003latent}.
In measuring document similarity within LDA, the \textit{Jensen-Shannon distance} is usually applied.

As a result of advances in GPU processing, deep learning models have been developed for learning high-dimensional continuous embeddings of corpora.
For example, \textit{Word2Vec} learns a representation for terms within a corpus by training a neural network that predicts the center word given its surroundings (\textit{Continuos BOW Model}) or vice versa by predicting the surrounding terms given the central word (\textit{Continuous Skip-gram Model})~\cite{mikolov2013efficient}.
\textit{Doc2Vec} extends the concept of Word2Vec for entire documents by training a neural network to predict the next word given a document together with words (\textit{Distributed Memory Model}) or a set of words given a document as input (\textit{Distributed BOW Model})~\cite{le2014distributed}.
The embeddings derived from Doc2Vec are learned in an iterative manner using back-propagation and allow for comparison using the cosine similarity.
We further consider \textit{Bidirectional Encoder Representations from Transformers} (BERT) - a deep learning model that is trained for two unsupervised NLP tasks and, as a result, generates high-dimensional representations for documents within a corpus~\cite{devlin2018bert}.
By grouping similar documents according to their latent representations, BERT is known for generating well-interpretable topics using a class-based tf-idf weighting~\cite{grootendorst2022bertopic}.

\subsection{Dimensionality Reductions}
As a result of the second stage, each document is represented in a latent space. To further project the documents to the two-dimensional plane, we apply a DR.
Thereby, we focus on DRs that are commonly used for text spatializations~\cite{acstrds2023-evaluation-tm-dr}: \textit{Metric Multidimensional Scaling} (MDS), \textit{Self-Organizing Maps} (SOMs), \textit{t-distributed Stochastic Neighbor Embedding} (t-SNE), and \textit{Uniform Manifold Approximation} (UMAP).
Furthermore, t-SNE and UMAP are probably the most popular DRs among practitioners and also show the best results in terms of accuracy in previous studies~\cite{Espadoto2021Toward,acstrds2023-evaluation-tm-dr}.
Although many more DRs exist, we limit our considerations to these four for capacity reasons.

MDS operates on a dissimilarity matrix of a dataset, aiming to generate a lower-dimensional representation where pairwise Euclidean distances reflect the entries in the dissimilarity matrix~\cite{cox2008multidimensional}.
The positions of data points are iteratively computed by optimizing a stress function.
The number of iterations constitutes a hyperparameter of the model.

SOMs constitute a class of fully-connected two-layered neural networks where second-layer neurons are organized on a two-dimensional grid, whose width and height are determined by hyperparameters~\cite{kohonen1997som}.
During training, input vectors activate neurons whose weight vectors are most similar to the input.
The neuron of the highest activation determines the position within the grid.
Weight adjustments during training minimize quantization errors, i.e., differences between input vectors and their best matching unit.
For computational efficiency, we utilized \textit{Principal Component Analysis} (PCA) to derive a lower-dimensional representation, that still captures 95\% of the dataset's variance~\cite{jolliffe2005principal}.

t-SNE is a DR aimed at preserving local structures within a dataset~\cite{van2008visualizing}.
It operates by modeling a Gaussian distribution centered around each data point in the high-dimensional space, where the perplexity hyperparameter regulates the effective number of neighbors considered.
The objective of t-SNE is to maintain neighborhood relationships in the low-dimensional representation using a t-distribution.
Its iterative optimization process minimizes a stress function, which evaluates the dissimilarity between overall similarity scores derived from the respective distributions and the \textit{Kullback-Leibler Divergence}.

UMAP was developed as an alternative to t-SNE to address its limitations, such as the difficulty in interpreting distances between clusters~\cite{mcinnes2020umap}.
While conceptually similar to t-SNE, UMAP optimizes a stress function based on \textit{Cross-Entropy} instead of Kullback-Leibler divergence.
UMAP offers two hyperparameters: the number of neighbors, balancing local and global structures, and the minimal distance, regulating the proximity of data points in the two-dimensional layout.

\subsection{Comparison}

\begin{table}
	\caption{Metrics used in our study. We did not specify an optimum for rotation, e.g., 0, as the rotation can be carried out as a postprocessing step when comparing two layouts. All metrics are invariant under rotation.}%
	\vspace{-1.0\baselineskip}
	%
%\tiny
\footnotesize%
\centering%
\setlength{\tabcolsep}{4.0pt}%
\renewcommand{\arraystretch}{1.00}%
\begin{tabular}{rccc}
	\toprule
	\multicolumn{1}{r}{\textbf{Metric}} & \multicolumn{1}{c}{\textbf{Abbr.}} & \multicolumn{1}{c}{\textbf{Range}} &\multicolumn{1}{c}{\textbf{Optimum}} \\ \midrule
	%GitHub Projects & \numprint[MiB]{2024.5} & 653 &  \numprint{52635} & \numprint{405117} & 8 \\
	%Ecommerce & \numprint[MB]{1783.9} & \numprint{50123} & \numprint{4665} & 4 \\
	Trustworthiness & $\alpha_T$ & [0,1] & 1\\
	Continuity & $\alpha_C$ & [0,1] & 1\\
	Mean Relative Rank Errors & $\alpha_{MM}$, $\alpha_{MF}$ & [0,1] & 1\\
	Local Continuity Meta-Criterion & $\alpha_{LC}$ & [0,1] & 1\\
	Label Preservation & $\alpha_{LP}$ & [0,1] & 1\\[0.6ex]
	Pearson's Correlation & $\beta_{PC}$ & [-1,1] & 1\\
	Spearman's Rank Correlation & $\beta_{SC}$ & [-1,1] & 1\\
	Cluster Ordering & $\beta_{CO}$ & [-1,1] & 1\\[0.6ex]
	Abs. Diff Distance Consistency & $\gamma_{DC}$ & [0,1] & 0\\[0.6ex]
	Rotation from Procrustes Analysis & $\theta_{PA}$ & [-180\textdegree,180\textdegree] & $\bot$\\
	\bottomrule
\end{tabular}%
    \label{tab:metrics}
\end{table}
Our analysis of stability requires a notion of similarity between scatterplots.
From the study of the related work, we derived three different types of metrics for quantifying scatterplot similarity: (1) latent representations learned from neural networks, (2) perceptual similarity features, and (3) features that capture selected aspects, e.g., neighborhood preservation.
We do not consider latent representations due to their lack of interpretability.
Additionally, we omit scagnostics, as they do not align well with human judgment~\cite{DBLP:conf/chi/PandeyKFBB16,Wang2020RobustScagnostics}.
Instead, we opted for metrics that quantify selected aspects of similarity, grouping them into local and global similarity, as well as cluster separation. To this end, we have adapted existing accuracy metrics, i.e., metrics that quantify the preservation of local and global structures of high-dimensional data in a low-dimensional representation~\cite{aupetit2007visualizing}.
We provide an overview in \Cref{tab:metrics}.

For quantifying local similarity, i.e., the preservation of neighborhoods, we consider six metrics.
The \textit{Trustworthiness} measure $\alpha_T$ is often used to quantify the accuracy of DRs~\cite{venna2006visualizing}.
It measures the pointwise-percentage of the k-Nearest-Neighbors (kNN) in the low-dimensional representation that also belong to the kNN in the high-dimensional representation, weighted by ranks and averaged over all points.
Similarly, the \textit{Continuity} $\alpha_C$ measures the proportion of points in the high-dimensional representation belonging to the kNN in the low-dimensional representation~\cite{venna2006visualizing}.
Both trustworthiness and continuity depend only on the pairwise dissimilarities of the points but not on the positions of the points themselves.
Therefore, both measures can also be used to compare two two-dimensional representations concerning the preservation of neighborhoods.
The same consideration holds for the \textit{Mean Relative Rank Errors} $\alpha_{MM}$ and $\alpha_{MF}$, which are related to trustworthiness and continuity but with slightly different weightings~\cite{lee2009quality}, and the \textit{Local Continuity Meta-Criterion} $\alpha_{LC}$, which measures the pointwise intersection between the kNN of a point in the two scatterplots averaged over all points~\cite{chen2009local}.
We further integrate the \textit{Label Preservation} metric $\alpha_{LP}$, which is similar to the $\alpha_{LC}$ but only considers the categories and not the positions.
In any case, the local metrics require the specification of $k$, i.e., the number of neighbors considered.
We choose $k=7$ to be aligned with previous studies~\cite{Espadoto2021Toward,vernier2020quantitative,vernier2021guided,acsds2024-topic-model-influence,acstrds2023-evaluation-tm-dr}.
The preservation of neighborhoods between two scatterplots is highly relevant for the visualization, since close data points are assumed to be similar according to the \textit{Gestalt principles}~\cite{ware2019information}.

To quantify global similarity, we use three metrics.
The \textit{Spearman Rank Correlation} $\beta_{SC}$~\cite{sidney1957nonparametric} and the \textit{Pearson Correlation} $\beta_{PC}$~\cite{Geng2005SupervisedDR} are derived from the \textit{Shephard Diagram} from the two scatterplots.
The Shephard diagram is a two-dimensional scatterplot whose points represent pairwise distances between two points in the first and second scatterplot~\cite{Joia2011LocalAffine}.
In the case of a perfect match, the Shephard diagram would thus be a subset of a straight line through the origin.
The similarity to the straight line is quantified by the two metrics.
We further developed the \textit{Cluster Ordering} metric $\beta_{CO}$, which relates the arrangement of categories between two scatterplots.
The metric is given by the Pearson correlation of the pairwise distances between the categories centers; i.e., it relies on two graphs derived from the scatterplots similar to measures from the \textit{Graph-based Family}~\cite{motta2015graph}.
All three metrics for the global similarity have a bounded value range. 
We do not consider metrics with unbounded range, e.g., \textit{Normalized Stress}~\cite{kruskal1964multidimensional} or \textit{Procrustes Distance}~\cite{goldberg2009local}, as it is not evident how to normalize these measures across several text corpora and scales.

For our third set of metrics, we assess the efficacy of discerning given categories.
Building upon the findings of Sedlmair and Aupetit, we use the \textit{Distance Consistency}~\cite{sedlmair2015data}.
This metric evaluates the proportion of points within the projected two-dimensional space, where the associated category center, defined as the mean of all points in that category, coincides with its nearest category center with respect to the Euclidean distance~\cite{sips2009selecting}.
To quantify the similarity between two scatterplots concerning class separation, we use the absolute difference $\gamma_{DC}$ between their distance consistencies.

These ten similarity metrics are used in our sensitivity analysis as follows:
To analyze the stability concerning input data, we compare scatterplots that differ only in the jitter amount, but with the other configurations fixed (\textit{ceteris paribus}).
To analyze the stability concerning hyperparameters, we pair scatterplots that differ in one hyperparameter setting with consecutive values.
To analyze stability concerning randomness, the scatterplots differ only in the defined random seed.
The selection process is illustrated in \Cref{fig:comparisons}.
Our data processing pipeline results in three tabular datasets, where each entry is given by a pair of scatterplots and similarity measures from the ten metrics.
As a post-processing step, one scatterplot can be rotated according to the rotation derived from Procrustes analysis.
The angle is determined to minimize the Procrustes distance, i.e., the squared pairwise distances, between two scatterplots~\cite{kendall1989survey}.
Before applying Procrustes analysis, we center both scatterplots such that the angle refers to rotation around the view center.
We include the rotation to the tabular datasets.

\begin{table}[t]
	\caption{Range for the hyperparameters considered in our experiments. Each configuration for one DR is combined with a dataset and TM.}
	\vspace{-0.7\baselineskip}
	%
%\tiny
\footnotesize
\centering
\setlength{\tabcolsep}{8.0pt}%
\renewcommand{\arraystretch}{1.00}%
\begin{tabular}{rrr}
\toprule
\multicolumn{1}{c}{\textbf{DR}} & \multicolumn{1}{c}{\textbf{Parameter Name}} & \multicolumn{1}{c}{\textbf{Values}} \\ \midrule   
MDS & \texttt{max\_iter} & 100--300 step size 50 \\[0.6ex]
SOM & \texttt{m} & 5--30 step size 5 \\
SOM & \texttt{n} & 5--30 step size 5 \\[0.6ex] %\midrule
UMAP & \texttt{min\_dist} & 0.0, 0.1, 0.25, 0.5, 0.8, 0.99 \\
UMAP & \texttt{n\_neighbors} & 2, 5, 10, 20, 50, 100, 200 \\[0.6ex] %\midrule
t-SNE & \texttt{learning\_rate} & 10, 28, 129, 359, 1000, auto\\
t-SNE & \texttt{n\_iter} & 1000, 2500, 5000, 10000\\
t-SNE & \texttt{perplexity} & 5--55 step size 10 \\ %\midrule
\bottomrule
\end{tabular}%
    \label{tab:parameters}
\end{table}

\subsection{Implementation}
Our implementation is designed for various embeddings, DRs, and metrics and can be extended in the future.
For each embedding and DR, we chose from specific implementations and hyperparameters.
The large number of layout configurations required the use of a computing cluster. 
The project is freely available on GitHub for reproducibility and reuse\footnotemark[1]; the generated data is linked as Zenodo archive\footnote{Zenodo archive \href{https://doi.org/10.5281/zenodo.12772898}{\textsc{doi}:10.5281/zenodo.12772898}}.

\subsubsection{Software Dependencies}
The implementation is based on Python 3.10 and depends on actively maintained libraries that are popular among practitioners for the embeddings and DRs.
Our text preprocessing pipeline relies on \textit{NLTK} (3.7) and \textit{spaCy} (3.4.3) for lemmatization.
For topic modeling and Doc2Vec, we use \textit{Gensim} (4.2.0).
The pretrained BERT models are provided by the \textit{Sentence Transformer} library (2.2.2).
For t-SNE and MDS, we use the implementation provided by \textit{scikit-learn} (1.2.1); for UMAP, we use \textit{umap-learn} (0.5.3); and for SOMs, we utilize the \textit{sparse-som} library (0.6.1)~\cite{Melka2019}.
For the similarity metrics, we adopted the approaches and implementations by Atzberger and Cech et al.~\cite{acstrds2023-evaluation-tm-dr} and \textit{ZADU}~\cite{Jeon2023Zadu}.

\subsubsection{Hyperparameter Settings}

We selected values for the hyperparameters of the DRs following the documentation of the respective library and the original papers.
The value ranges for the hyperparameters for the DRs are specified in \Cref{tab:parameters}.
For each corpus-embedding combination, we used a fixed configuration for the embedding, i.e., we did not iterate over the embedding's hyperparameters.
When applying TMs, we set the number of topics $K$ to the number of categories $k$ in the case of the 20 Newsgroup corpus, $K = 2k$ for the Seven Categories corpus, and $K=3k$ for the Lyrics corpus, since the latter two have relatively few categories.
We followed best practices in choosing the TM's hyperparameters and inspected the topics of each trained TM to ensure interpretable topics~\cite{Wallach2009RethinkingMatter,Wallach2009EvaluationModels}. 
The extracted topics are provided in the supplemental material.
For BERT, we chose the two pre-trained models, \texttt{all-mpnet-base-v2} and \texttt{all-distilroberta-v1}, as they have shown the highest scores for the sentence embedding task\footnote{\href{https://www.sbert.net/docs/pretrained\_models.html}{sbert.net/docs/pretrained\_models.html}}.
Doc2Vec requires the specification of the embedding dimension and the number of iterations.
Following best practices, the trained models have an accuracy of above 95\% to predict the nearest neighbors of the inferred documents to be themselves among the three nearest neighbors\footnote{\href{https://radimrehurek.com/gensim/auto\_examples/tutorials/run\_doc2vec\_lee.html\#sphx-glr-auto-examples-tutorials-run-doc2vec-lee-py}{radimrehurek.com/gensim/auto\_examples/tutorials/run\_doc2vec}}.

\begin{figure}[t]
    %\vspace{-5pt}
    \input{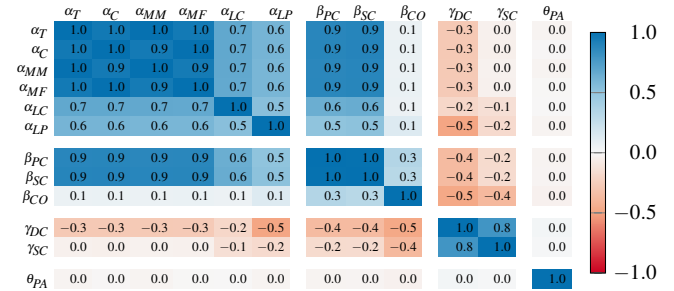}%
    \vspace{-0.7\baselineskip}
    \caption{Heatmap showing the pairwise correlations between the similarity metrics using a diverging color scheme. We additionally show the correlation with the \textit{Silhouette Coefficient}, which is another cluster separation metric. Metrics that correlate nearly perfect, i.e., $\alpha_T, \alpha_C, \alpha_{MM}, \alpha_{MF}$ as well as $\beta_{PC}, \beta_{SC}$ are considered as one metric by taking their averages. Note: the local and global similarity measures show a negative correlation to the class separation measures, as they have opposite optimums.}%
    \label{fig:Correlation}%
\end{figure}

\begin{figure*}[t]
    \includegraphics[]{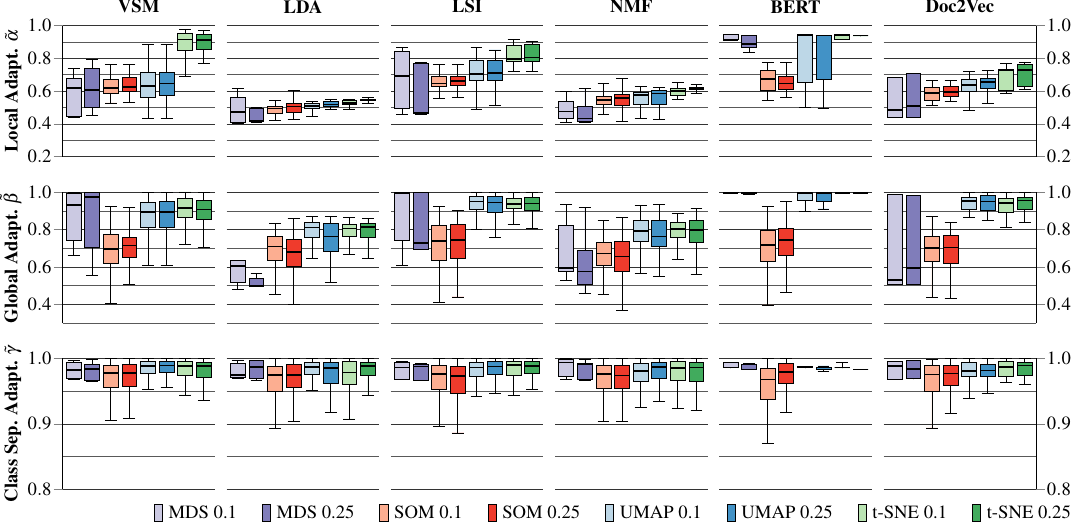}%
    \vspace{-1.6\baselineskip}
    \caption{Results of the first experiment to quantify the stability concerning changes to the input data.
        The hue of each bar indicates the DR, and the intensity indicates the amount of jitter applied to the DTM.
        The metrics $\tilde{\alpha}$, $\tilde{\beta}$, and $\tilde{\gamma}$ quantify how well the layout algorithm adapts to changes to the DTM, with 1 being optimal.
        The visualization indicates that BERT, in combination with t-SNE, best reflects changes to the DTM concerning $\tilde{\alpha}$ and $\tilde{\beta}$, resulting in improvements compared to the VSM.
        Note: The vertical axis ranges differ between the three metrics $\tilde{\alpha}$, $\tilde{\beta}$, and $\tilde{\gamma}$.}
    \label{fig:boxplots-jittering}
\end{figure*}

\subsubsection{Computational Cluster}

The setup of the computational cluster is similar to the previous study of Atzberger and Cech et al.~\cite{acstrds2023-evaluation-tm-dr}.
Specific to this study, we used jobs with a RAM limit of \numprint[GB]{40} and used up to 29 nodes to optimize for a higher throughput of jobs.
In total, our experiments had a run time of over \numprint{49000} CPU hours.
From the targeted \numprint{40860} layouts, the cluster could compute \numprint{38941} layouts (\numprint[\%]{95.3}).
The unsuccessful computations can be accounted to timeouts after \numprint[hours]{30} (120 layouts) and general abortions due to exceeding RAM, planned downtime, unplanned downtime, file system errors, etc. (1799 layouts).

\section{Data Analysis \& Results}
\label{section: setup}

In total, our three datasets that resulted from comparing pairs of scatterplots resulted in \numprint{42817} data points.
Due to interruptions in the computational cluster, not all layouts were created successfully.
On average, each dataset contains \numprint{14272} pairs of scatterplots.
We first analyze the correlation between the metrics to derive an aggregated metric for local similarity, global similarity, and similarity concerning cluster separation.
The distributions of data points are inspected using a series of boxplots, which allows us to analyze the three stability aspects covered in our study.
In two binary tests, we analyze the effect of the tf-idf weighting scheme and the application of the DR on the topic representations.

\subsection{Correlation of Similarity Metrics}

To derive a higher-level overview of the similarity aspects, we aggregate several metrics to represent the three similarity aspects, e.g., by taking an average~\cite{Espadoto2021Toward,vernier2020quantitative,vernier2021guided,Morariu2023PredictingUserPreferences}.
However, relying on the average carries the risk of overweighting within one aspect, e.g., in case most metrics are strongly correlated.
Alternatively, taking a weighted average based on pairwise correlations counteracts this, but conversely, it might outweigh a single metric that is not correlated to the others at all~\cite{acsds2024-topic-model-influence}.
\Cref{fig:Correlation} shows the pairwise correlations of the ten similarity metrics, as well as the \textit{Silhouette Coefficient} as a further class separation metric and the rotation that is derived from Procrustes analysis.
To derive the correlations, we randomly selected 3000 pairs of scatterplots, equally distributed over the corpora.

Regarding local similarity, there is a strong positive correlation between all metrics; the first four metrics correlate nearly perfectly.
Therefore, we averaged the first four metrics. 
Thus, we define the aggregated local similarity metric $\alpha$ as 
\begin{align}
    \alpha = \dfrac{1}{3} \Bigg(
                \dfrac{\alpha_{T} + \alpha_{C} + \alpha_{MM} + \alpha_{MF}}{4} + \alpha_{LC} + \alpha_{LP} \Bigg).
\end{align}

Regarding global similarity, the Spearman and Pearson correlation measures correlate perfectly.
Therefore, we consider their average and pair it with the cluster ordering metric.
Furthermore, we apply an affine transformation to all three metrics, to translate their value ranges to [0,1].
Thus, we define the global similarity measure $\beta$ as
\begin{align}
    \beta = \dfrac{1}{2}\Bigg( \dfrac{0.5\cdot (\beta_{PC}+1) + 0.5\cdot (\beta_{SC}+1)}{2} + \dfrac{1}{2}(\beta_{CO}+1) \Bigg).
\end{align}

To quantify changes to the class separability, we use the absolute difference between the distance consistencies between two scatterplots.
As $\gamma_{DC}$ has its optimum at 0, we define the metric $\gamma$ as:
\begin{align}
    \gamma = 1- \gamma_{DC}
\end{align}

\begin{figure*}[t]
    \includegraphics[]{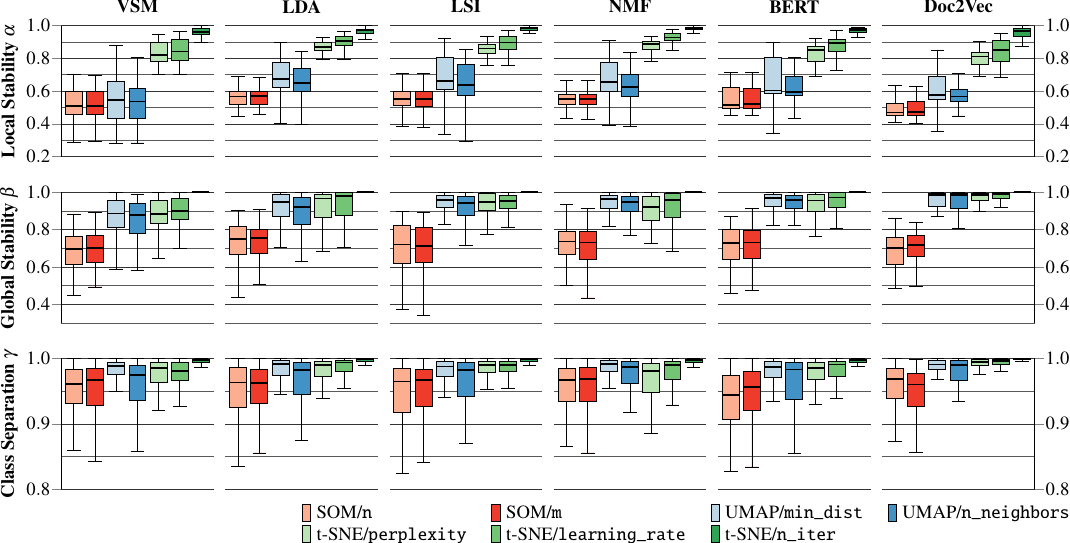}%
    \vspace{-1.6\baselineskip}
    \caption{Results of the second experiment to quantify the stability concerning hyperparameters.
    The hue of each bar indicates the DR and the intensity indicates a specific hyperparameter that is varied.
    LDA, LSI, and NMF, in combination with t-SNE, show the highest stability concerning changes to the hyperparameters.
    Note: The vertical axis ranges differ between the three metrics $\alpha$, $\beta$, and $\gamma$.}
    \label{fig:boxplots-hyperparameters}
\end{figure*}

\subsection{Stability Concerning Input Data}
Starting from the three similarity metrics $\alpha$, $\beta$, and $\gamma$, we first analyze the stability of text layout algorithms concerning changes to the input data.
For this, we applied jittering on the DTM and compared scatterplots that result from the same layout algorithm with the same hyperparameter configurations.
A stable layout algorithm should reflect the changes to the DTM in the layout.
Given a layout algorithm $\Phi$, we quantify the adaptability of the layout algorithm to changes in the local structures of the DTM using the following metric:
\begin{align}
    \tilde{\alpha} = 1 - \vert\alpha(\text{DTM}, \text{DTM}^{\text{jitter}}) - \alpha(\Phi(\text{DTM}), \Phi(\text{DTM}^{\text{jitter}}))\vert,
\end{align}
where $\alpha(\text{DTM}, \text{DTM}^{\text{jitter}})$ denotes the local similarity between the DTM and its jittered variant, and $\alpha(\Phi(\text{DTM}), \Phi(\text{DTM}^{\text{jitter}}))$ denotes the local similarity between the two scatterplots $\Phi(\text{DTM})$ and $\Phi(\text{DTM}^{\text{jitter}})$.
Analogously, we define $\tilde{\beta}$ to quantify how well the layout algorithm adapts to changes in the DTM concerning global structures and $\tilde{\gamma}$ how well changes concerning cluster separation are captured.
In the case of a value of 1, the scatterplots reflect changes to $\alpha$, $\beta$, and $\gamma$ perfectly.
The results are shown in \Cref{fig:boxplots-jittering}.

t-SNE best reflects changes to local structures of the DTM, as it shows the highest values for $\tilde{\alpha}$.
Furthermore, in most cases, t-SNE and UMAP adapt best to changes in the DTM's global structures measured by $\tilde{\beta}$.
All DRs show appropriate changes regarding cluster separation measured by $\tilde{\gamma}$, but SOMs show the largest range across all embeddings.

Changes to the DTM are reflected poorly in the case of layouts based on LDA.
We assume that LDA extracts similar topics for both the DTM and its jittered variant and, therefore, represents both very similarly in the latent space. 
Thus, the dissimilarity of the DTM and its jittered variant is not reflected in the latent space, which means that the resulting scatterplots do not depict the desired change either after applying a DR.
NMF shows the same effect as LDA.
Doc2Vec also leads to decreases in $\tilde{\alpha}$ but reflects global changes well in the case of t-SNE and UMAP.
BERT and LSI reflect changes in global structures well and improve the results shown by the VSM.
However, only BERT shows improvements concerning $\tilde{\alpha}$ compared to the VSM for all four DRs.
In summary, BERT in combination with t-SNE best reflects changes to the input data.

\subsection{Stability Concerning Hyperparameters}
In the second experiment, we analyze the stability of the layout algorithms concerning small changes to the hyperparameters.
\Cref{fig:boxplots-hyperparameters} shows the pairwise similarities between pairs of scatterplots that differ in consecutive values in exactly one hyperparameter.
We omit MDS since, in 100 percent of all cases, MDS has converged after 200 iterations, i.e., the pairwise similarities are one.
Among the remaining three DRs, SOMs are the most sensitive to changes in their hyperparameters (referring to their median) concerning $\alpha$, $\beta$, and $\gamma$. 
The hyperparameters height and width have nearly the same impact due to the symmetry of the grid structure of the SOM layout.
UMAP shows a significant sensitivity concerning local similarity $\alpha$ but is much more stable regarding global similarity $\beta$.
Changes to the minimum distance do not affect the cluster separation $\gamma$ strongly.
t-SNE shows the highest scores for all three metrics.
From the small interquartile distance of the boxes displaying the number of iterations, we deduce that the algorithm converges quite early.
The perplexity parameter has the most significant impact on t-SNE layouts' stability.
But still, it is more stable than SOMs and UMAP.
Our results contradict the widespread statement that t-SNE creates unstable layouts.
However, the metrics used ignore perceptual differences due to rotation, as they are invariant under rotation.

Each embedding improves the stability concerning changes in the hyperparameters.
The improvements through LDA, LSI, and NMF are comparable in each case, i.e., we see no clear benefit in choosing one over the other.
However, the TMs outperform BERT and Doc2Vec in terms of $\alpha$.
The dimensions of the latent spaces are significantly lower for the three TMs than in the case of Doc2Vec (50) and the two BERT models (768 for both embeddings). 
We assume that higher dimensionality can result in stronger distortions due to the DRs.
To summarize, t-SNE in combination with a topic model shows the most stable behavior concerning changes to its input parameters.

\subsection{Stability Concerning Randomness}
\begin{figure}[t]
    \includegraphics[]{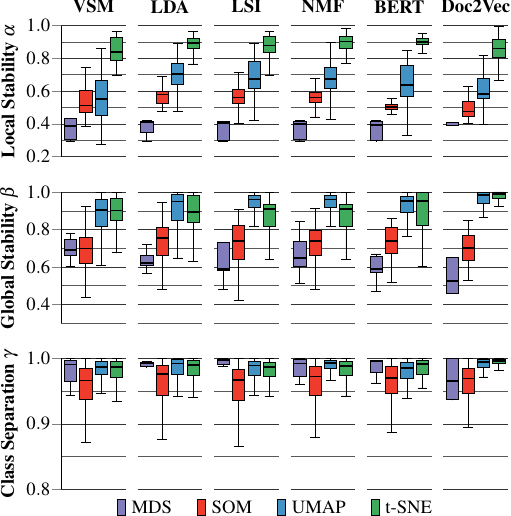}%
    \vspace{-0.6\baselineskip}
    \caption{Results of the third experiment to quantify the stability concerning randomness.
    The color of each bar indicates the DR underlying the layout algorithm.
    Overall, LDA in combination with t-SNE shows the best results. Note: The vertical axis ranges differ between the three metrics $\alpha$, $\beta$, and $\gamma$.}
    \label{fig:boxplots-randomness}
\end{figure}

By specifying a random seed of the DR, the layouts are reproducible across multiple runs. 
Since different initializations might result in different local optima, two scatterplots derived from the same layout algorithm can thus differ.
\Cref{fig:boxplots-randomness} shows the values for the similarity metrics between scatterplots that only differ in the random seed.

The results show that differences in the random seed can be very considerable, e.g., in the case of MDS concerning $\alpha$.
t-SNE shows the most stable behavior concerning $\alpha$ by far, i.e., the neighborhoods are represented similarly regardless of the selected seed.
UMAP, followed by t-SNE, achieves the highest stability concerning global structures.
MDS, UMAP, and t-SNE show stable behavior concerning $\gamma$, whereas SOMs can lead to larger changes.

For UMAP and t-SNE, embeddings can further increase the stability concerning $\alpha$.
Furthermore, embeddings improve the global stability in the case of UMAP.
The stability concerning class separation shows high values across all embeddings.
Overall, we favor t-SNE due to the strong dominance concerning $\alpha$.
In combination with LDA, the local stability can further be improved.

\subsection{Binary Tests}

In our previous experiments, we aggregated different layout algorithms that share the same embedding and DR.
However, the VSM, LSI, and NMF can be applied to either the DTM or its tf-idf weighted variant.
Atzberger and Cech et al. have empirically shown that the tf-idf weighting improves layout accuracy and perception~\cite{acstrds2023-evaluation-tm-dr}.
We analyze whether this observation also transfers to stability using a binary test.
For this, we compare the aggregated stability metrics $\tilde{\alpha}$, $\tilde{\beta}$, and $\tilde{\gamma}$ in the case of stability concerning input data, and $\alpha$, $\beta$, and $\gamma$ in the case of the other two stability aspects of two pairs of scatterplots $(\Phi_1(\text{DTM}), \Phi_2(\text{DTM}))$ and $(\Phi_1(\text{DTM}^{\text{tf-idf}}), \Phi_2(\text{DTM}^{\text{tf-idf}}))$, where $\Phi_1$ and $\Phi_2$ are layout algorithms with specified hyperparameters that were compared in the respective experiment.
For all possible combinations of such pairs of tuples, we determine the occurrences $n_{\alpha}$, $n_{\beta}$, and $n_{\gamma}$ in which the tf-idf weighted variant shows larger values.
Assuming that the tf-idf weighting has no impact on the stability, the values $n_{\alpha}$, $n_{\beta}$, and $n_{\gamma}$ are distributed according to a binomial distribution with probability 0.5.
Given this distribution, we determine the probability of our observations for $n_{\alpha}$, $n_{\beta}$, and $n_{\gamma}$.
In the case, of a small probability, we reject the hypothesis, since a large value would be unlikely in the case of probability 0.5.
Our results are summarized in \Cref{tab:binary_test_tfidf}.
A hypothesis is usually rejected if the probability is smaller than 0.05.
In this case, the improvement made using the tf-idf scheme is statistically validated.

\begin{table}[t]
	\caption{Results of the binary test for the null hypothesis \enquote{The tf-idf weighting scheme does not improve the stability concerning input data (S1), hyperparameters (S2), or randomness (S3).}}%
	\vspace{-1\baselineskip}
\centering
\tiny
\setlength{\tabcolsep}{4.0pt}%
\renewcommand{\arraystretch}{1.0}%
\begin{tabular}{ccSSSSSSSSS}
	\toprule%\addlinespace[1em]
	\multicolumn{2}{c}{} &
	\multicolumn{3}{c}{\scriptsize\bfseries S1}&
	\multicolumn{3}{c}{\scriptsize\bfseries S2}&
	\multicolumn{3}{c}{\scriptsize\bfseries S3}\\
	%\multicolumn{1}{c}{\bfseries D(1)}
	%&\\
	\cmidrule(rl){3-5}\cmidrule(rl){6-8}\cmidrule(rl){9-11}
	%\cmidrule(rl){5-6}\cmidrule(rl){7-8}\cmidrule(rl){9-10}\cmidrule(rl){11-12}
	{\footnotesize\bfseries TM}&
	{\footnotesize\bfseries DR}&
	\multicolumn{1}{c}{\scriptsize$\tilde{\alpha}$} & \multicolumn{1}{c}{\scriptsize$\tilde{\beta}$} & \multicolumn{1}{c}{\scriptsize$\tilde{\gamma}$} & \multicolumn{1}{c}{\scriptsize$\alpha$} & \multicolumn{1}{c}{\scriptsize$\beta$} & \multicolumn{1}{c}{\scriptsize$\gamma$} & \multicolumn{1}{c}{\scriptsize$\alpha$} & \multicolumn{1}{c}{\scriptsize$\beta$} & \multicolumn{1}{c}{\scriptsize$\gamma$} \\
	\midrule
	\parbox[t]{2mm}{\multirow{4}{*}{\rotatebox[origin=c]{90}{\scriptsize VSM}}} & MDS & 0.50 & 0.03 & 0.94 & 0.95 & 0.95 & 0.95 & 0.94 & 0.30 & 0.94 \\
	 & SOM & 0.45 & 0.99 & 0.66 & 0.24 & 0.99 & 0.47 & 0.42 & 1.00 & 0.84\\
	 & t-SNE & 0.00 & 0.00 & 0.01 & 0.00 & 0.99 & 0.03 & 0.00 & 1.00 & 0.23\\
	 & UMAP & 0.00 & 0.00 & 0.05 & 0.00 & 0.16 & 0.24 & 0.00 & 0.09 & 0.39\\
	\midrule
	\parbox[t]{2mm}{\multirow{4}{*}{\rotatebox[origin=c]{90}{\scriptsize LSI}}} & MDS & 0.65 & 0.65 & 0.01 & 1.00 & 1.00 & 1.00 & 0.05 & 0.98 & 0.99\\
	 & SOM & 0.00 & 1.00 & 0.99 & 0.00 & 0.99 & 0.99 & 0.00 & 0.99 & 0.99\\
	 & t-SNE & 0.00 & 1.00 & 0.05 & 0.00 & 0.99 & 0.99 & 1.00 & 0.02 & 1.00\\
	 & UMAP & 0.00 & 0.99 & 0.99 & 0.00 & 0.99 & 0.99 & 0.00 & 1.00 & 1.00\\
	\midrule
	\parbox[t]{2mm}{\multirow{4}{*}{\rotatebox[origin=c]{90}{\scriptsize NMF}}} & MDS & 0.05 & 0.01 & 0.21 & 1.00 & 1.00 & 1.00 & 0.05 & 0.00 & 0.57\\
	 & SOM & 0.07 & 0.05 & 0.92 & 0.00 & 0.99 & 0.99 & 0.00 & 1.00 & 0.77\\
	 & t-SNE & 1.00 & 0.99 & 0.00 & 0.00 & 1.00 & 0.99 & 1.0 & 1.0 & 1.0\\
	 & UMAP & 0.00 & 0.00 & 0.99 & 0.00 & 0.99 & 0.99 & 0.00 & 0.97 & 1.00\\
	\bottomrule
\end{tabular}
\pgfplotsset{colormap/Blues-4}
\adjustbox{valign=m}{
\hspace*{-3.2cm}
\begin{tikzpicture}[inner sep=0pt,outer sep=0pt]
\begin{axis}[
    hide axis,
    scale only axis,
    height=0cm,
    width=0cm,
    /pgfplots/colormap name=Blues-4,
    colorbar horizontal,
    point meta min=0,
    point meta max=1,
    colorbar style={
	width=3.5cm,
	height=0.2cm,
	rotate=90,
	xtick style={draw=gray},
	xtick={1,0.5,0},
	xticklabels={$0.0$,$0.5$,$1.0$},
	xticklabel style={
	    font=\scriptsize,
	    xshift=3pt,
	    yshift=-0.2ex,
	    anchor=west,
	},
	typeset ticklabels with strut,
	at={(0.0,0.0)},anchor=south,
    }]
    \addplot [draw=none] coordinates {(0,0) (1,1)};
\end{axis}
\end{tikzpicture}
}
    \label{tab:binary_test_tfidf}
\end{table}

\begin{table}[t]
	\caption{Results of the binary test for the null hypothesis \enquote{\Cref{eq: convex combination} does not improve the stability concerning input data (S1), hyperparameters (S2), or randomness (S3).}}%
	\vspace{-1\baselineskip}
\centering
\tiny
\setlength{\tabcolsep}{4.0pt}%
\renewcommand{\arraystretch}{1.0}%
\begin{tabular}{ccSSSSSSSSS}
	\toprule%\addlinespace[1em]
	\multicolumn{2}{c}{} &
	\multicolumn{3}{c}{\scriptsize\bfseries S1}&
	\multicolumn{3}{c}{\scriptsize\bfseries S2}&
	\multicolumn{3}{c}{\scriptsize\bfseries S3}\\
	%\multicolumn{1}{c}{\bfseries D(1)}
	%&\\
	\cmidrule(rl){3-5}\cmidrule(rl){6-8}\cmidrule(rl){9-11}
	%\cmidrule(rl){5-6}\cmidrule(rl){7-8}\cmidrule(rl){9-10}\cmidrule(rl){11-12}
	{\footnotesize\bfseries TM}&
	{\footnotesize\bfseries DR}&
	 \multicolumn{1}{c}{\scriptsize$\tilde{\alpha}$} & \multicolumn{1}{c}{\scriptsize$\tilde{\beta}$} & \multicolumn{1}{c}{\scriptsize$\tilde{\gamma}$} & \multicolumn{1}{c}{\scriptsize$\alpha$} & \multicolumn{1}{c}{\scriptsize$\beta$} & \multicolumn{1}{c}{\scriptsize$\gamma$} & \multicolumn{1}{c}{\scriptsize$\alpha$} & \multicolumn{1}{c}{\scriptsize$\beta$} & \multicolumn{1}{c}{\scriptsize$\gamma$} \\
	\midrule
	\parbox[t]{2mm}{\multirow{4}{*}{\rotatebox[origin=c]{90}{\scriptsize LDA}}} & MDS & 0.92 & 0.42 & 0.15 & 0.69 & 0.69 & 0.69 & 1.00 & 0.07 & 0.00 \\
	 & SOM & 0.00 & 0.00 & 0.12 & 0.13 & 0.00 & 0.09 & 0.16 & 0.00 & 0.01\\
	 & t-SNE & 0.00 & 0.00 & 0.00 & 0.00 & 0.00 & 0.00 & 0.00 & 0.00 & 0.00\\
	 & UMAP & 0.00 & 0.00 & 0.03 & 0.00 & 0.00 & 0.30 & 0.03 & 0.00 & 0.10\\
	\midrule
	\parbox[t]{2mm}{\multirow{4}{*}{\rotatebox[origin=c]{90}{\scriptsize LSI}}} & MDS & 1.00 & 0.50 & 1.00 & 1.00 & 1.00 & 1.00 & 1.00 & 0.70 & 1.00\\
	 & SOM & 0.75 & 0.11 & 0.87 & 0.05 & 0.41 & 0.76 & 0.01 & 0.04 & 0.78\\
	 & t-SNE & 0.09 & 0.86 & 0.09 & 0.26 & 0.48 & 0.70 & 0.89 & 0.77 & 0.06\\
	 & UMAP & 0.85 & 0.55 & 0.45 & 0.55 & 0.83 & 0.50 & 0.01 & 0.02 & 0.27\\
	\midrule
	\parbox[t]{2mm}{\multirow{4}{*}{\rotatebox[origin=c]{90}{\scriptsize NMF}}} & MDS & 0.29 & 0.01 & 0.29 & 1.00 & 1.00 & 1.00 & 0.15 & 0.00 & 0.15\\
	 & SOM & 0.00 & 0.07 & 0.99 & 0.00 & 0.93 & 0.99 & 0.00 & 0.99 & 0.81\\
	 & t-SNE & 0.99 & 0.01 & 0.63 & 0.00 & 1.00 & 0.99 & 0.48 & 1.00 & 1.00\\
	 & UMAP & 0.00 & 0.00 & 0.99 & 0.00 & 0.99 & 0.99 & 0.00 & 0.94 & 0.94\\
	\bottomrule
\end{tabular}
\pgfplotsset{colormap/Blues-4}
\adjustbox{valign=m}{
\hspace*{-3.2cm}
\begin{tikzpicture}[inner sep=0pt,outer sep=0pt]
\begin{axis}[
    hide axis,
    scale only axis,
    height=0cm,
    width=0cm,
    /pgfplots/colormap name=Blues-4,
    colorbar horizontal,
    point meta min=0,
    point meta max=1,
    colorbar style={
	width=3.5cm,
	height=0.2cm,
	rotate=90,
	xtick style={draw=gray},
	xtick={1,0.5,0},
	xticklabels={$0.0$,$0.5$,$1.0$},
	xticklabel style={
	    font=\scriptsize,
	    xshift=3pt,
	    yshift=-0.2ex,
	    anchor=west,
	},
	typeset ticklabels with strut,
	at={(0.0,0.0)},anchor=south,
    }]
    \addplot [draw=none] coordinates {(0,0) (1,1)};
\end{axis}
\end{tikzpicture}
}
    \label{tab:binary_test_convexcombination}
\end{table}

In the case of the VSM, we see that the tf-idf weighting in combination with t-SNE and UMAP improves all three stability aspects in terms of preserving local structures.
Furthermore, for both t-SNE and UMAP, it supports the layout algorithm to adapt to changes in the input data.
Also, for LSI, the tf-idf weighting scheme improves stability regarding changes in the input data and hyperparameters in combination with t-SNE and UMAP concerning local structures.
The combination of NMF and UMAP profits from the tf-idf weighting in all three stability aspects concerning the preservation of local structures.
We want to point out that a high probability does not indicate the validity of the null hypothesis.
For example, the high probability for LSI in combination with t-SNE concerning stability concerning randomness might be due to the small value range of the similarity measures, as shown in \Cref{fig:boxplots-randomness}.

Our second binary test concerns the input for the DR.
Most visualization approaches apply the DR to the document representations of the corpus in the latent space~\cite{acstrds2023-evaluation-tm-dr}.
In the case of a TM, the components of the document representations describe the importance of a topic within the document.
However, the topics might have different similarities, which are not taken into account when applying the DR directly.
Atzberger et al. proposed an alternative by applying the DR on the topics themselves and deriving the document position in the two-dimensional plane as a linear combination, i.e., the position $\bar{d}$ of a document $d$ is given by
\begin{align} \label{eq: convex combination}
    \bar{d} = \sum \limits_{j=1}^{K} \theta_j \bar{\phi}_j,
\end{align}
where $\theta = (\theta_1, \dots, \theta_K)$ denotes the topic representation of $d$, and $\bar{\phi}_1, \dots, \bar{\phi}_K$ denotes the positions of the topics after application of a DR~\cite{atzberger2021softwareforest}.
Analogously, we compute the probabilities for the null hypothesis that \Cref{eq: convex combination} does not improve the three stability aspects.
The results are shown in \Cref{tab:binary_test_convexcombination}.

LDA, in combination with t-SNE, benefits from \Cref{eq: convex combination} across all three stability aspects in terms of local and global structures, as well as class separation.
In the case of LSI and NMF, the results are not that obvious.
We suspect that in the case of LSI and NMF, the extracted topics are \enquote{more orthogonal} to each other since these two TMs rely on eigenvectors.
In the case of LDA, this is not the case, and therefore, the dissimilarities between the topics differ more, which is emphasized in \Cref{eq: convex combination}.

\section{Discussion}
\label{section: discussion}
From the results of our evaluation, we see certain combinations of text embeddings and DRs that are particularly suitable for generating stable two-dimensional layouts for text corpora.
We use these observations together with the findings of Atzberger and Cech et al.~\cite{acstrds2023-evaluation-tm-dr,acsds2024-topic-model-influence} to derive guidelines for the effective combination of text embeddings and DRs.
However, our analysis, as well as the guidelines derived from it, are subject to threats to validity.

\subsection{Main Findings}
A visualization designer has to make numerous design decisions when creating text spatializations.
A fundamental one is whether the DR should be applied to the DTM or the embedding of the corpus in a latent space.
From the three sensitivity analyses, we have shown that text embeddings can improve all three aspects of stability.
As such, we conclude that:
\begin{description}
    \item[\textbf{G1}] We recommend using a text embedding to increase the stability concerning input data, hyperparameters, and randomness.
\end{description}
Furthermore the three sensitivity analyses show, that depending on the specific stability aspect, the text embeddings differ in their performance.
Therefore, depending on the stability aspect that is to be optimized, we recommend the following embeddings:
\begin{description}
    \item[\textbf{G2-S1}] We recommend BERT when optimizing for stability concerning input data.
    \item[\textbf{G2-S2}] We recommend LDA, LSI, and NMF when optimizing for stability concerning hyperparameters.
    \item[\textbf{G2-S3}] We recommend LDA when optimizing for stability concerning randomness.
\end{description}
When applying LDA, we derive from the corresponding binary tests: 
\begin{description}
    \item[\textbf{G3}] We recommend applying an aggregation according to \Cref{eq: convex combination} when applying LDA.
\end{description}
When not optimizing for one specific stability aspect, but for all three, we further consider the weaknesses of each embedding. 
Since BERT shows poor results concerning accuracy~\cite{acstrds2023-evaluation-tm-dr}, and LDA is very sensitive to changes in the input parameters, we conclude:
\begin{description}
    \item[\textbf{G4}] We recommend using LSI as text embedding when optimizing for all three stability aspects.
\end{description}
In all experiments, t-SNE showed the best results, especially in terms of the preservation of local structures. We therefore conclude: 
\begin{description}
    \item[\textbf{G5}] We recommend using t-SNE as the dimensionality reduction.
\end{description}
In particular, these guidelines align with the recommendations concerning accuracy from previous studies~\cite{acstrds2023-evaluation-tm-dr,acsds2024-topic-model-influence}.

\subsection{Threats to Validity}
Our results depend on specific choices and have thus threats to validity.
We see two major areas: (1) the sampling used in the data measurement step and (2) errors in the implementation and execution.

For the data measurement, we selected text corpora, text embeddings, DRs, and similarity metrics.
In any of the four categories, we had to select a subset among many possibilities.
Our results rely on three text corpora.
A priori, it is unclear to what extent the patterns in the boxplot visualizations depend on the specific corpora.
We added the boxplot visualizations for each corpus to the supplemental material.
In any case, we see that the patterns in the individual boxplot visualizations are similar to the aggregated view. 
Therefore, our argumentation from section 4 is still valid individually.
We assume additional corpora will not affect our results, particularly the derived guidelines. 
For each corpus-embedding combination, we fixed the hyperparameters of the embedding algorithm following best practices.
We furthermore inspected the resulting topics of the TMs and compared them to the given categories to ensure that the model is of high quality.
It is unclear if guideline G2 transfers to the case where more variants of each embedding are evaluated.
Nevertheless, one of our main findings, that embeddings can improve the stability concerning changes to the input data, the hyperparameters, and the random seed, was validated in our experiments by using embeddings following best practices.
Lastly, even our similarity metrics required the specification of hyperparameters, such as the number of points $k$ to be considered as nearest neighbors. 
In choosing $k = 7$, we followed previous studies and emphasized the metrics to capture local similarity.

For the entire data processing pipeline, we used actively maintained libraries that are widely used among practitioners.
However, we can not guarantee that these libraries have no bugs and do not change their behavior across releases.
Furthermore, our implementations, e.g., the similarity metrics, could carry defects.
We addressed this by using only code reviewed by at least one co-author and pair-programming sessions.
Finally, we provide our entire code as a GitHub repository to allow for transparency.
Due to errors in the cluster, some layouts were not computed.
Therefore, some scatterplot pairings could not be evaluated.
It is unclear to what extent the missing values would affect the results.

\section{Conclusions \& Future Work}
\label{section: conclusions}
Many visualizations for text corpora rely on a two-dimensional scatterplot that is derived from applying a text embedding and a subsequent DR.
Since changes to the layout require cognitive effort by the user, the stability of a layout algorithm needs to be considered by the visualization designer.
In this study, we analyzed the stability of text layout algorithms concerning changes to the input data, hyperparameters, and randomness.
For this, we measured the preservation of local and global structures and cluster separation between a large set of scatterplots that were derived from systematically iterating over the layout algorithms and the hyperparameters of the underlying DRs.
Based on a correlation analysis of the similarity metrics, we aggregated them into three metrics to quantify the similarity concerning local structures, global structures, and class separation.
Based on a detailed statistical analysis of the results, we analyzed the impact of the embedding algorithms and the DRs concerning the different stability aspects.
We discussed our findings and derived guidelines for the effective use of text embeddings and DRs to generate text spatializations.
Our work aims to address uncertainties when applying text embeddings and DRs for the visualization of text corpora.
Furthermore, we hope practitioners and researchers consider our guidelines when applying latent embeddings and DRs.
To draw a \enquote{big picture}, we further see possible applications of our evaluation setup -- particularly the metrics -- for selecting DRs for exploring different embeddings, e.g., internal representations of neural network approaches.
The findings from such experiments could be integrated into visualization approaches that aim to help users explain high-dimensional embeddings~\cite{wang2023wizmap,Boggust2022EmbeddingComparator,Sivaraman2022Emblaze}.

We see different directions for future work, e.g., by extending our experiments to address the major threats to validity.
We plan to evaluate different configurations for each embedding to derive fine-granular insights into their impact on layout stability.
Our approach for evaluating stability can further be adapted to quantify the stability of layouts of time-dependent text corpora.
Such experiments would require additional, time-dependent embeddings and DRs.
Furthermore, it would be interesting to measure the similarity of scatterplots by using additional metrics, e.g., Aupetit and Sedlmair presented a large set of class separation measures~\cite{Aupetit2016Sepme}.
Quantifying stability using measures that best align with human similarity perception is particularly relevant concerning the preservation of the mental map.
We plan to conduct a user study to verify to what extent our guidelines improve text spatializations for concrete analysis tasks.
The results from such a user study might also lead to a deeper understanding of how humans perceive similarity in scatterplots.
Furthermore, it would be interesting to see whether our findings for text embeddings might generalize to other types of data, including multidimensional tabular datasets.
While our measurements do not directly address the stability of DRs in multidimensional tabular data, the latent embeddings used in our study could analogously represent tabular data, given their lower dimensionality and sparsity ratio.
Finally, our pursuit of guidelines for text specializations motivates their generalization to multimodal corpora, i.e., sets of documents that include other modalities such as images.
In particular, this would require new kinds of embeddings and a more complex description of latent embeddings across several data domains~\cite{Huang2023UseofEmbeddings}.

\acknowledgments{
We thank the reviewers for their valuable feedback.
This work was partially funded by the Federal Ministry of Education and Research, Germany through grant 01IS22062 and project 16KN086467 funded by the Federal Ministry for Economic Affairs and Climate Action of Germany.
The work of Tobias Schreck was partially funded by the Austrian Research Promotion Agency (FFG) within the framework of the flagship project ICT of the Future PRESENT, grant FO999899544.
}

\bibliographystyle{abbrv-doi-hyperref-narrow}

\bibliography{main-shortened}

\appendix



\section*{Supplementary material}

# A Large-Scale Sensitivity Analysis on Latent Embeddings and Dimensionality Reductions for Text Spatializations
## Supplemental Material

Daniel Atzberger ⓘ, Tim Cech ⓘ, Willy Scheibel ⓘ,
Jürgen Döllner ⓘ, Michael Behrisch ⓘ, and Tobias Schreck ⓘ

## A  TOPICS

Table 1 – Table 21 show the top ten words for selected topics of all considered Topic Models.

## B  INDIVIDUAL BOXPLOTS

Figure 1 show the results of the first experiment to quantify the stability concerning changes to the input data on the three individual corpora.

Figure 2 show the results of the second experiment to quantify the stability concerning hyperparameters on the three individual corpora.

Figure 3 show the results of the third experiment to quantify the stability concerning randomness on the three individual corpora.

Table 1: Top ten words for the first ten selected topics for the LDA model for the 20 Newsgroup corpus.

| 0 | 1 | 2 | 3 | 4 | 5 | 6 | 7 | 8 | 9 |
|---|---|---|---|---|---|---|---|---|---|
| line | organization | god | would | health | key | car | gun | file | window |
| organization | line | believe | write | line | write | organization | people | line | line |
| would | post | write | line | medical | would | line | right | organization | organization |
| post | host | say | organization | organization | chip | write | would | drive | write |
| get | write | one | go | disease | organization | article | government | include | problem |
| write | de | know | get | report | clipper | university | state | system | one |
| host | university | belief | say | center | encryption | price | say | available | get |
| know | research | article | one | research | make | would | write | sale | driver |
| need | version | people | know | year | line | post | law | new | drive |
| like | article | science | post | child | article | get | one | send | work |

Table 2: Top ten words for the first ten selected topics for the `all_distilroberta_v1` model for the 20 Newsgroup corpus.

| 0 | 1 | 2 | 3 | 4 | 5 | 6 | 7 | 8 | 9 |
|---|---|---|---|---|---|---|---|---|---|
| car | di | team | tiff | key | science | food | cool | space | photography |
| host | um | game | philosophical | encryption | methodology | patient | tower | launch | het |
| window | mu | player | significance | clipper | fulk | disease | nuclear | orbit | van |
| want | pu | play | spec | chip | lady | medical | plant | satellite | object |
| right | ah | season | gripe | privacy | rational | doctor | water | moon | energy |
| distribution | mi | win | cartwright | escrow | scientific | health | steam | mission | picture |
| thank | ax | hockey | martin | security | hypothesis | dyer | uranium | henry | ether |
| thing | te | league | hitchhiker | government | homeopathy | cancer | reactor | shuttle | photograph |
| computer | mo | baseball | nova | algorithm | theory | bank | fossil | lunar | field |
| problem | eu | score | universe | secure | experiment | pain | temperature | rocket | plate |

Table 3: Top ten words for the first ten selected topics for the `all_mpnet_base_v2` model for the 20 Newsgroup corpus.

| 0 | 1 | 2 | 3 | 4 | 5 | 6 | 7 | 8 | 9 |
|---|---|---|---|---|---|---|---|---|---|
| god | team | comet | wire | drug | eye | di | seizure | patient | food |
| window | game | orbit | outlet | legalization | dominance | um | corn | disease | sensitivity |
| want | player | temporary | ground | legalize | handedness | mu | cereal | doctor | superstition |
| work | play | radii | wiring | hypocrisy | correction | pu | dyer | medical | restaurant |
| problem | hockey | perijove | neutral | marijuana | cornea | ah | disorder | treatment | taste |
| come | season | kelvin | circuit | cocaine | dominant | mi | diet | science | eat |
| version | win | shoemaker | conductor | alcohol | eyedness | ax | epilepsy | pain | reaction |
| need | league | levy | breaker | grader | almanac | te | food | infection | effect |
| host | baseball | observatory | prong | wod | lens | eu | bu | health | substance |
| file | score | orbital | grounding | war | contact | ne | eat | yeast | flavor |

Table 4: Top ten words for the first ten selected topics for the LSI model for the 20 Newsgroup corpus.

| 0 | 1 | 2 | 3 | 4 | 5 | 6 | 7 | 8 | 9 |
|---|---|---|---|---|---|---|---|---|---|
| one | file | file | wire | gun | wire | team | drive | image | di |
| file | say | entry | entry | entry | file | hockey | disk | god | um |
| say | go | system | file | file | team | god | hard | drive | mu |
| go | entry | image | wiring | program | hockey | image | controller | window | ah |
| get | program | output | image | image | wiring | game | card | atheist | mi |
| would | image | available | ground | control | image | league | anonymous | go | pu |
| know | know | say | circuit | firearm | game | go | bit | color | eu |
| people | people | gun | outlet | law | entry | season | support | gif | ne |
| make | would | window | gun | new | league | anonymous | server | format | ax |
| well | think | version | say | say | year | new | information | system | pa |

Table 5: Top ten words for the first ten selected topics for the LSI (tf-idf) model for the 20 Newsgroup corpus.

| 0 | 1 | 2 | 3 | 4 | 5 | 6 | 7 | 8 | 9 |
|---|---|---|---|---|---|---|---|---|---|
| would | window | key | key | drive | god | card | drive | car | space |
| get | file | game | god | window | gun | driver | gun | window | file |
| one | card | team | chip | file | car | drive | window | bike | henry |
| people | god | chip | game | ide | card | file | file | gun | gun |
| say | drive | clipper | clipper | car | key | gun | ide | key | sale |
| window | driver | encryption | team | controller | game | video | disk | god | car |
| know | disk | player | encryption | card | drive | disk | sale | driver | zoo |
| go | people | win | window | disk | ide | mouse | mail | mouse | god |
| think | video | play | player | program | bike | window | god | space | window |
| like | thank | hockey | escrow | win | chip | diamond | controller | ride | launch |

Table 6: Top ten words for the first ten selected topics for the NMF model for the 20 Newsgroup corpus.

| 0 | 1 | 2 | 3 | 4 | 5 | 6 | 7 | 8 | 9 |
|---|---|---|---|---|---|---|---|---|---|
| write | file | say | team | entry | di | space | one | image | window |
| line | gun | go | hockey | program | um | launch | see | available | atheist |
| article | output | know | game | output | mu | satellite | wire | format | may |
| organization | line | come | year | build | ah | day | time | version | argument |
| get | firearm | tell | league | line | available | law | come | also | exist |
| post | bill | something | season | char | pu | make | two | include | believe |
| host | check | see | new | year | mi | mission | wiring | system | example |
| university | write | think | play | section | eu | prophecy | even | display | atheism |
| would | control | take | win | remark | ne | orbit | ground | program | question |
| know | handgun | happen | player | check | pa | man | find | color | well |

Table 7: Top ten words for the first ten selected topics for the NMF (tf-idf) model for the 20 Newsgroup corpus.

| 0 | 1 | 2 | 3 | 4 | 5 | 6 | 7 | 8 | 9 |
|---|---|---|---|---|---|---|---|---|---|
| clipper | gun | mouse | drive | car | state | mac | god | monitor | bike |
| encryption | firearm | stratus | disk | convex | right | chip | believe | sun | ride |
| chip | morality | science | ide | engine | government | code | people | program | dod |
| government | atheist | computer | hard | radar | people | graphic | religion | bony | dog |
| key | crime | year | floppy | dealer | country | book | faith | jake | motorcycle |
| escrow | weapon | rise | controller | detector | kill | group | say | win | helmet |
| privacy | criminal | rocket | boot | speed | bank | audio | life | thank | rider |
| space | law | well | apple | light | turkey | source | church | de | wave |
| algorithm | moral | colorado | problem | buy | drug | want | atheist | prism | get |
| technology | evidence | child | mac | law | attack | bit | belief | help | honda |

Table 8: Top ten words for the first ten selected topics for the `all_distilroberte_v1` model for the Lyrics corpus.

| 0 | 1 | 2 | 3 |
|---|---|---|---|
| na | la | na | big |
| tit | sha | anna | real |
| shove | ay | ganja | money |
| anybody | plastic | fool | hold |
| stand | crocodile | pilot | home |
| send | mas | whip | gon |
| bottle | es | ready | turn |
| dream | ah | burn | hand |
| blah | thrill | future | new |
| lose | speak | sympathy | talk |

Table 9: Top ten words for the first ten selected topics for the `all_mpnet_base_v2` model for the Lyrics corpus.

| 0 | 1 | 2 |
|---|---|---|
| treasure | real | te |
| kodak | play | tu |
| stand | hold | ne |
| smile | new | bom |
| amazing | hey | se |
| fly | home | en |
| birthday | hand | ce |
| lose | turn | toi |
| save | bad | si |
| grateful | long | es |

Table 10: Top ten words for the first ten selected topics for the LDA model for the Lyrics corpus.

| 0 | 1 | 2 | 3 | 4 | 5 | 6 | 7 | 8 | 9 |
|---|---|---|---|---|---|---|---|---|---|
| la | da | know | yeah | hey | get | na | get | whoa | yeah |
| go | burn | go | get | long | know | ah | like | oh | go |
| dance | cry | come | like | tonight | like | ready | bitch | big | get |
| get | like | time | oh | alright | go | ha | oh | de | like |
| run | gun | well | ya | go | say | hot | back | bang | know |
| hand | go | love | back | hello | come | rain | shake | daddy | say |
| ah | play | night | go | fire | one | like | yeah | eh | want |
| girl | get | never | love | dream | man | oh | dog | shoot | tell |
| like | high | life | know | like | might | boy | know | get | see |
| away | walk | might | take | oh | want | go | say | aye | time |

Table 11: Top ten words for the first ten selected topics for the LSI model for the Lyrics corpus.

| 0 | 1 | 2 | 3 | 4 | 5 | 6 | 7 | 8 | 9 |
|---|---|---|---|---|---|---|---|---|---|
| get | oh | yeah | na | na | like | go | la | know | bitch |
| like | get | get | love | love | get | love | know | la | like |
| yeah | bitch | like | get | get | bitch | get | go | go | know |
| know | like | know | oh | bitch | go | let | love | love | go |
| go | love | go | bitch | go | love | like | ah | ah | love |
| oh | yeah | love | know | oh | huh | bitch | bitch | like | baby |
| bitch | na | say | go | know | feel | know | like | get | ass |
| love | baby | see | baby | like | run | see | let | see | get |
| say | em | never | like | yeah | ass | say | see | tell | feel |
| make | ah | one | one | baby | look | come | oh | say | la |

Table 12: Top ten words for the first ten selected topics for the LSI (tf-idf) model for the Lyrics corpus.

| 0 | 1 | 2 | 3 | 4 | 5 | 6 | 7 | 8 | 9 |
|---|---|---|---|---|---|---|---|---|---|
| yeah | bitch | oh | la | ah | yeah | love | na | baby | hey |
| oh | yeah | yeah | ah | la | oh | na | love | na | baby |
| love | love | ah | oh | oh | bitch | baby | bitch | hey | love |
| get | oh | baby | da | hey | love | yeah | hey | girl | cry |
| bitch | gon | la | na | ha | ah | want | oh | bitch | world |
| baby | get | na | yeah | na | baby | bitch | la | love | night |
| know | huh | hey | girl | baby | girl | girl | little | yeah | blue |
| want | heart | whoa | ha | gon | big | lord | heart | little | long |
| go | money | ill | hey | bitch | em | away | yeah | heart | home |
| let | ill | na | sha | alright | ass | home | want | home | light |

Table 13: Top ten words for the first ten selected topics for the NMF model for the Lyrics corpus.

| 0 | 1 | 2 | 3 | 4 | 5 | 6 | 7 | 8 | 9 |
|---|---|---|---|---|---|---|---|---|---|
| back | know | see | want | like | oh | get | yeah | love | get |
| yeah | go | come | make | bitch | yeah | go | gang | hey | yeah |
| say | love | em | money | feel | love | like | come | say | bitch |
| time | yeah | make | go | look | like | baby | ah | never | money |
| see | let | hand | run | make | get | say | baby | one | ass |
| tell | baby | let | gang | say | take | let | like | make | never |
| know | time | fly | well | run | way | tell | hard | ah | big |
| make | good | might | time | em | baby | cause | one | right | huh |
| well | tell | boy | like | huh | come | need | take | man | give |
| let | see | also | blow | pull | feel | know | drink | need | new |

Table 14: Top ten words for the first ten selected topics for the NMF (tf-idf) model for the Lyrics corpus.

| 0 | 1 | 2 | 3 | 4 | 5 | 6 | 7 | 8 | 9 |
|---|---|---|---|---|---|---|---|---|---|
| bitch | la | woo | oh | na | love | yeah | back | baby | away |
| big | die | bitch | huh | man | baby | gon | want | ass | life |
| gon | ill | huh | nobody | get | girl | la | come | money | run |
| bad | man | yo | really | em | heart | baby | let | get | still |
| ass | cry | gang | get | different | good | crazy | take | real | feel |
| get | wait | em | whoa | bitch | oh | bitch | friend | yeah | lose |
| ho | walk | get | let | hit | ill | get | around | bang | change |
| pull | little | fly | talk | hoe | tonight | girl | ride | look | ya |
| bout | hear | keep | tell | shoot | night | let | play | bitch | way |
| em | life | dick | never | buy | always | boy | get | talk | mind |

Table 15: Top ten words for the first ten selected topics for the `all_distilroberte_v1` model for the Seven Categories corpus.

| 0 | 1 | 2 | 3 | 4 | 5 | 6 | 7 | 8 | 9 |
|---|---|---|---|---|---|---|---|---|---|
| vector | fraction | cache | united | plant | wave | share | share | debenture | cash |
| quantity | lesson | bit | soviet | organism | harmonic | allotment | allotment | redemption | book |
| measurement | decimal | address | war | cell | frequency | capital | capital | issue | purchase |
| dimension | algebra | register | president | animal | amplitude | equity | company | discount | date |
| significant | denominator | processor | party | specie | oscillation | money | money | premium | cheque |
| error | triangle | memory | government | water | displacement | application | equity | redeemable | transaction |
| statement | variable | computer | union | tissue | periodic | company | application | application | sale |
| ratio | solution | datum | political | transport | motion | receive | premium | asset | journal |
| mass | math | page | military | biology | sound | premium | receive | reserve | bank |
| heat | review | clock | country | blood | string | forfeit | reserve | journal | petty |

Table 16: Top ten words for the first ten selected topics for the `all_mpnet_base_v2` model for the Seven Categories corpus.

| 0 | 1 | 2 | 3 | 4 | 5 | 6 | 7 | 8 | 9 |
|---|---|---|---|---|---|---|---|---|---|
| organism | error | cash | cache | fraction | register | nucleus | united | hormone | share |
| animal | trial | book | miss | lesson | computer | neutron | soviet | gland | allotment |
| cell | account | purchase | page | decimal | bit | decay | war | pregnancy | capital |
| protein | balance | cheque | block | algebra | branch | nuclear | president | sperm | money |
| acid | credit | date | virtual | denominator | clock | proton | government | reproductive | company |
| waste | debit | sale | hierarchy | triangle | performance | nucleon | party | female | application |
| plant | sale | transaction | memory | variable | pipeline | reactor | union | stimulate | equity |
| bacteria | post | journal | associative | solution | datum | reaction | political | uterus | receive |
| enzyme | suspense | bank | access | math | processor | bind | military | follicle | premium |
| microbe | rectify | petty | address | review | disk | fission | country | ovum | pay |

Table 17: Top ten words for the first ten selected topics for the LDA model for the Seven Categories corpus.

| 0 | 1 | 2 | 3 | 4 | 5 | 6 | 7 | 8 | 9 |
|---|---|---|---|---|---|---|---|---|---|
| plant | cash | specie | cash | pollen | war | number | issue | plant | share |
| organism | share | premium | payment | purchase | also | cache | biology | call | call |
| also | asset | root | transaction | force | force | rectify | account | figure | company |
| animal | call | number | date | muscle | new | bit | two | body | error |
| figure | statement | call | account | two | united | block | balance | process | journal |
| gene | bank | account | student | body | become | figure | also | organism | amount |
| call | blood | balance | university | war | soviet | one | cell | water | pay |
| one | account | figure | amount | one | year | time | call | also | time |
| share | membrane | seed | particular | soviet | state | two | one | form | balance |
| food | activity | per | civil | asset | president | write | figure | high | book |

Table 18: Top ten words for the first ten selected topics for the LSI model for the Seven Categories corpus.

| 0 | 1 | 2 | 3 | 4 | 5 | 6 | 7 | 8 | 9 |
|---|---|---|---|---|---|---|---|---|---|
| force | instruction | share | share | cell | cache | charge | debenture | number | number |
| two | memory | force | charge | plant | instruction | force | cash | force | wave |
| instruction | register | call | war | water | memory | instruction | force | wave | current |
| time | address | charge | call | force | miss | cache | share | debenture | bit |
| one | cache | capital | force | charge | block | motion | account | cash | time |
| call | force | application | energy | instruction | register | field | time | cell | force |
| also | bit | allotment | united | call | charge | electric | charge | share | light |
| number | war | energy | new | organism | page | body | wave | account | magnetic |
| memory | write | money | soviet | form | time | particle | book | charge | water |
| point | datum | amount | instruction | field | system | current | balance | time | instruction |

Table 19: Top ten words for the first ten selected topics for the LSI (tf-idf) model for the Seven Categories corpus.

| 0 | 1 | 2 | 3 | 4 | 5 | 6 | 7 | 8 | 9 |
|---|---|---|---|---|---|---|---|---|---|
| instruction | share | share | war | share | cache | cell | debenture | charge | fraction |
| register | instruction | debenture | charge | debenture | miss | cache | cash | motion | decimal |
| memory | register | allotment | soviet | cash | instruction | debenture | sale | wave | number |
| bit | debenture | instruction | magnetic | purchase | block | plant | account | magnetic | bit |
| address | memory | cash | united | account | register | cash | cell | particle | lesson |
| cache | cash | charge | party | sale | memory | charge | purchase | field | exponent |
| branch | bit | application | government | allotment | page | water | plant | electric | denominator |
| charge | address | money | field | balance | branch | organism | redemption | current | instruction |
| cycle | allotment | capital | president | redemption | charge | miss | error | velocity | multiply |
| clock | charge | force | electric | book | hierarchy | magnetic | issue | cell | review |

Table 20: Top ten words for the first ten selected topics for the NMF model for the Seven Categories corpus.

| 0 | 1 | 2 | 3 | 4 | 5 | 6 | 7 | 8 | 9 |
|---|---|---|---|---|---|---|---|---|---|
| water | charge | state | light | force | current | call | bit | motion | energy |
| surface | cache | united | number | particle | coil | share | number | soviet | register |
| potential | force | soviet | time | energy | magnetic | two | page | force | one |
| air | time | war | branch | two | axis | force | address | body | call |
| pressure | system | new | also | mass | loop | time | cache | union | write |
| plant | miss | government | party | motion | unit | cell | memory | bus | instruction |
| earth | memory | union | become | body | electron | issue | instruction | change | datum |
| cell | motion | party | example | axis | wire | wave | block | momentum | address |
| mass | body | become | wave | point | light | give | table | external | cycle |
| equation | work | attack | frequency | velocity | length | debenture | set | device | plant |

Table 21: Top ten words for the first ten selected topics for the NMF (tf-idf) model for the Seven Categories corpus.

| 0 | 1 | 2 | 3 | 4 | 5 | 6 | 7 | 8 | 9 |
|---|---|---|---|---|---|---|---|---|---|
| current | register | magnetic | electron | charge | light | soviet | cache | right | instruction |
| memory | pressure | speed | instruction | force | energy | error | computer | decimal | point |
| page | share | fraction | round | electric | plant | union | branch | percent | disk |
| coil | vector | equation | cycle | resistance | electron | party | practice | number | float |
| cash | energy | variable | wave | motion | wave | president | instruction | china | share |
| force | potential | pole | current | field | nucleus | state | velocity | war | system |
| asset | instruction | air | write | capacitor | radiation | account | motion | leg | control |
| mirror | war | water | clock | reactor | atom | communist | performance | black | charge |
| machine | datum | dipole | ray | current | cell | trial | processor | length | io |
| frequency | read | induce | settlement | direction | photon | republic | acceleration | lesson | memory |

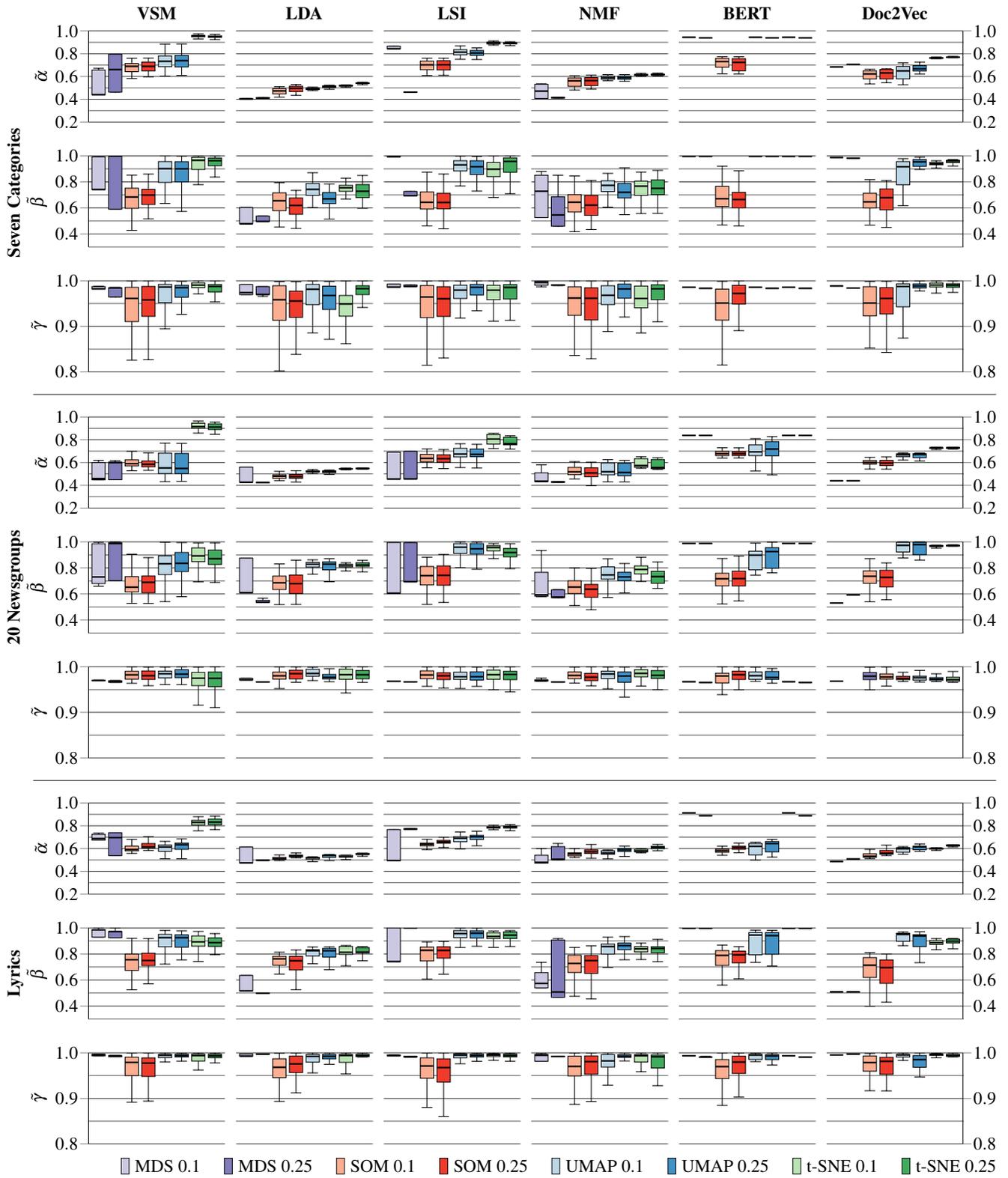

Figure 1: Results of the first experiment to quantify the stability concerning changes to the input data among the three datasets.

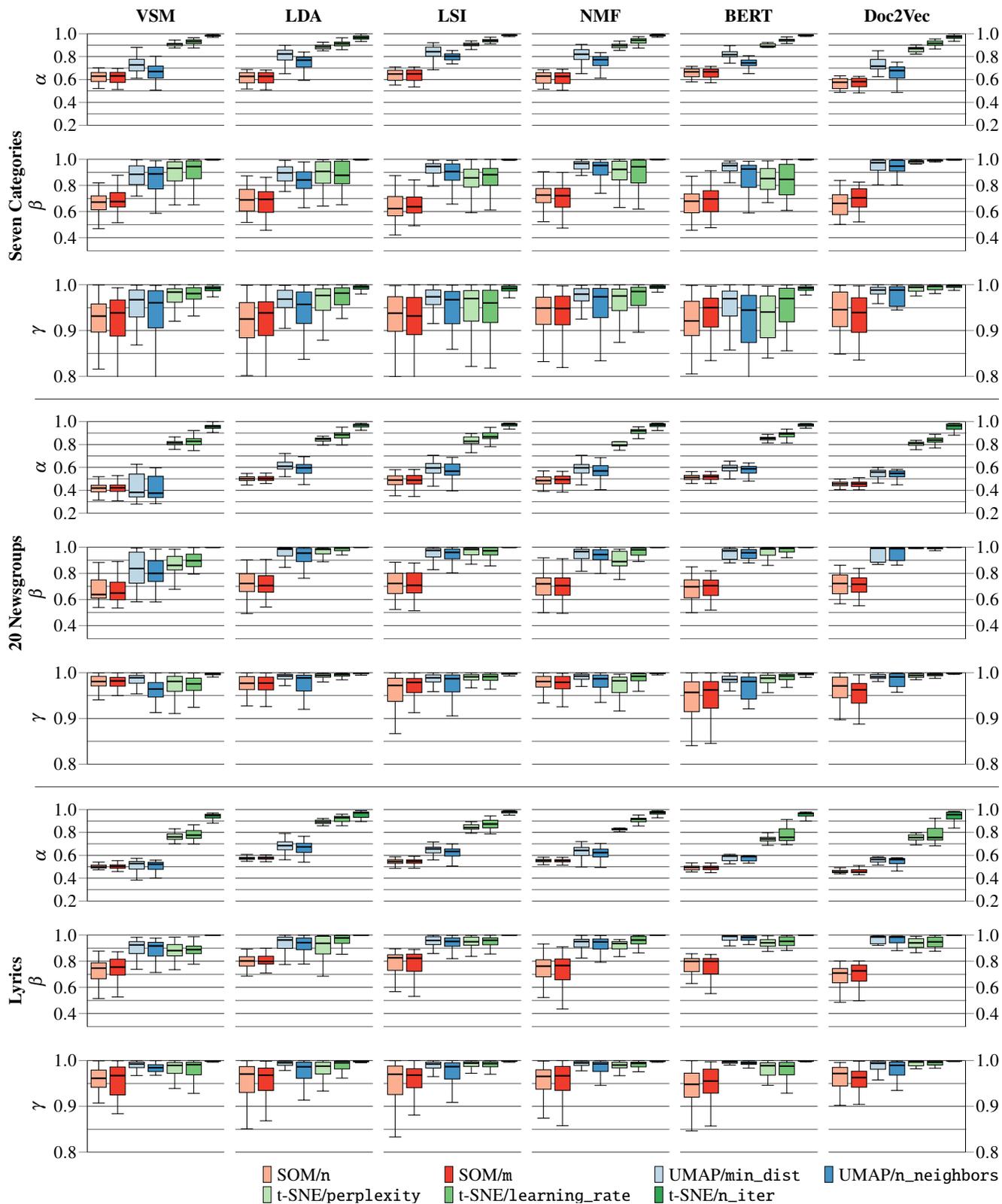

Figure 2: Results of the second experiment to quantify the stability concerning hyperparameters among the three datasets.

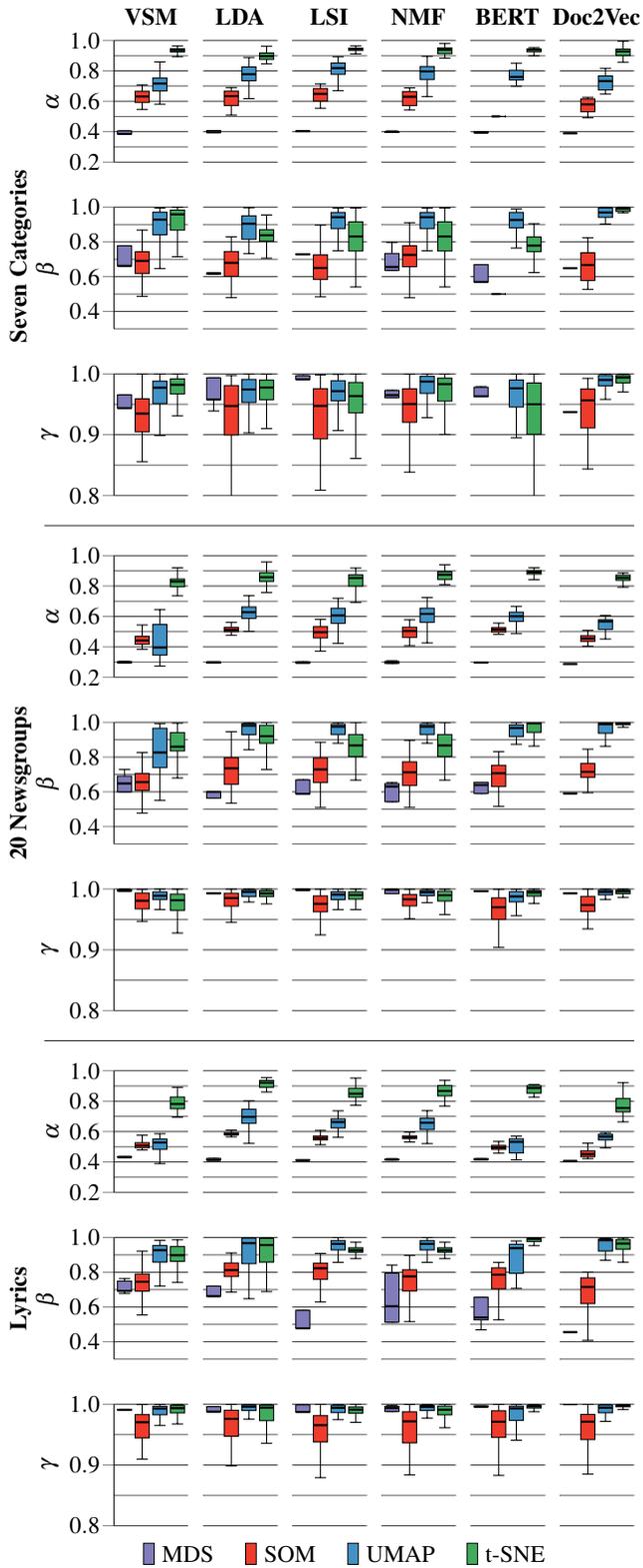

Figure 3: Results of the third experiment to quantify the stability concerning randomness among the three datasets.



\end{document}